%% file: SAPEOemo2017.tex
\documentclass[runningheads,a4paper]{llncs2e/llncs}

\usepackage{amssymb}
\usepackage{amsmath}
\usepackage{amsfonts}

\usepackage{graphicx}

\usepackage{url}
\urldef{\mailsa}\path|{vanessa.volz, guenter.rudolph}@tu-dortmund.de| 
\urldef{\mailsb}\path|boris.naujoks@th-koeln.de|    
\newcommand{\keywords}[1]{\par\addvspace\baselineskip
\noindent\keywordname\enspace\ignorespaces#1}

\usepackage[dvipsnames]{xcolor}
\usepackage{paralist}
\usepackage{listings}
\usepackage[autostyle=true]{csquotes}
\usepackage{dsfont}
\usepackage{tikz}
\usepackage[hidelinks]{hyperref}
\usepackage{caption}
\usepackage{subcaption}
\usepackage{adjustbox}

\usepackage[colorinlistoftodos]{todonotes}
\usepackage{algorithm}
\usepackage{algorithmic}
\floatname{algorithm}{Algorithm}

\newcommand{\epsi}{\ensuremath{\varepsilon}}

\begin{document}

\mainmatter  

\title{Surrogate-Assisted Partial Order-based Evolutionary Optimisation}

\titlerunning{Surrogate-Assisted Partial Order-Based Evolutionary Optimisation}

%
%
\author{Vanessa Volz\inst{1} \and G\"unter Rudolph\inst{1} \and Boris Naujoks\inst{2}}
\authorrunning{Volz, Rudolph, Naujoks}

\institute{TU Dortmund University\\
\mailsa \medskip
\and 
TH K\"oln - University of Applied Sciences\\
\mailsb}

%
%

\toctitle{Surrogate-Assisted Partial-order based Evolutionary Optimisation}
\tocauthor{Volz, Rudolph, Naujoks}
\maketitle

\input{src}

\bibliographystyle{splncs03}
\bibliography{sapeo} 

\end{document}

%% file: src.tex
\begin{abstract}
	In this paper, we propose a novel approach (SAPEO) to support the survival selection process in multi-objective evolutionary algorithms with surrogate models - it dynamically chooses individuals to evaluate exactly based on the model uncertainty and the distinctness of the population. We introduce variants that differ in terms of the risk they allow when doing survival selection. Here, the anytime performance of different SAPEO variants is evaluated in conjunction with an SMS-EMOA using the BBOB bi-objective benchmark. We compare the obtained results with the performance of the regular SMS-EMOA, as well as another surrogate-assisted approach. The results open up general questions about the applicability and required conditions for surrogate-assisted multi-objective evolutionary algorithms to be tackled in the future.
\keywords{partial order, multi-objective, surrogates, evolutionary algorithms, bbob}
\end{abstract}

\section{Introduction}
Surrogate model-assisted multi-objective evolutionary algorithms (SA-MOEAs) are a group of fairly recent but popular approaches\footnote{Workshop on the topic in 2016: http://samco.gforge.inria.fr/doku.php} to solve multi-objective problems with expensive fitness functions. Using surrogate model predictions of the function values instead of / to complement exact evaluations within an evolutionary algorithm (EA) can save computational time and in some cases make the problem tractable at all. A variety of SA-MOEAs have been proposed (see \cite{Jin2005} and section \ref{sec:samoea}), some of which use model predictions for a subset of the individuals (individual-based surrogate management).

Many MOEAs use the objective values of the individuals obtained through a function evaluation to sort and then select the best individuals in a population only. More formally, for the selection process in most MOEAs, a strict partial order of individuals is induced through non-dominated sorting. Selection is based on the obtained ranks and a total preorder created by a secondary criterion, if necessary. Assuming that individuals can confidently be distinguished based on surrogate model predictions, the individuals' exact objective values do not necessarily need to be known to make the correct selection decisions. In that case, no other part of the algorithm would be affected by trusting the predicted values, but the computational budget could be reduced. However, because of potential prediction errors, basing the selection process entirely on surrogate models can also lead the EA astray.

In this paper, we present a novel approach to integrate surrogate models and (multi-objective) evolutionary algorithms (dubbed SAPEO for \textbf{S}urrogate-\textbf{A}ssisted \textbf{P}artial Order-based \textbf{E}volutionary \textbf{O}ptimisation) that seeks to reduce the number of function evaluations without sacrificing too much solution quality. The idea is to choose the individuals for exact evaluation dynamically based on the model uncertainty and the distinctness of the population. Preliminary experiments on single-objective problems showed promising results, so in this paper, we investigate the approach in the currently sought-after multi-objective context. We also present different versions that control the risk of errors caused by using uncertain predictions over the actual values.

In the following, we describe our extensive analysis of the anytime performance of SAPEO using the BBOB-BIOBJ benchmark (refer to section \ref{sec:bbob}), focusing on use cases with low budgets to approximate applications with expensive functions. Our SAPEO implementation\footnote{code and visualisations available at: http://url.tu-dortmund.de/volz} uses Kriging \cite{Sacks1989} as a surrogate model and the SMS-EMOA \cite{Beume2007} as a MOEA. We further compare the algorithm and its variants to the underlying SMS-EMOA and a SA-MOEA approach called \textit{pre-selection} \cite{Emmerich} (SA-SMS in the following).

We specifically investigate if and under which conditions SAPEO outperforms the SMS-EMOA and SA-SMS in terms of the hypervolume indicator that all algorithms use to evaluate populations. Surprisingly, non of the surrogate-assisted algorithms can convincingly beat out the baseline SMS-EMOA, even for small function budgets. This result opens up questions about SA-MOEAs in general and about the necessary quality of the integrated surrogate models.

A potential explanation for this performance is the increased uncertainty of the surrogate model predictions when compared to the single-objective experiments. Thus, we analyse the effects of prediction uncertainty on the overall performance of the SA-MOEAs. In the future, the resulting insights could become important when (1) deciding whether using a surrogate model is beneficial on a given problem at all and (2) when choosing the sample size and further parameters for the model. This is especially crucial for multi- and many-objective problems, where learning an accurate surrogate model becomes increasingly expensive and thus renders analysing the trade-off between surrogate model computation and function evaluations critical.

In the following, we present related work in section 2 and introduce the proposed SAPEO algorithm in section 3. The benchmarking results and their discussion is found in section 4. Section 5 concludes the paper with a summary and open research questions.

\section{Background and Related work}

\subsection{Algorithm}
\label{sec:samoea}
Let $X_1, \dots, X_\lambda \in \mathbb{R}^n$ be a population and the corresponding fitness function $f: \mathbb{R}^n \rightarrow \mathbb{R}^d$. General concepts of multi-objective optimisation will not be discussed here. We will be referring to Pareto-dominance as $\preceq$, and to its weak and strong versions as $\precsim$ and $\prec$, respectively. We use the same notation to compare vectors in objective space $\mathbb{R}^d$, i.e. let $a, b \in \mathbb{R}^d$, then $a \preceq b \iff \forall {k \in \{1\dots d\}}: \, a_{k} \leq b_{k} \wedge \exists k \in \{1\dots d \}: a_k < b_k$.  

Surrogate-assisted multi-objective evolutionary optimisation is surveyed in \cite{Jin2005}, where several approaches for the integration of surrogates and MOEAS, or surrogate management strategies, are described. According to the survey, the selection approaches can generally be divided into individual-based, generation-based, and population-based strategies. Additionally, there is \textit{pre-selection}, where additional offspring are created and the required number of individuals is chosen using a (local) surrogate model before the usual survival selection is executed. The strategy in this paper is individual-based, like the established \textit{best strategy}, where only a number of individuals with the best predicted values are evaluated.

We propose a way to frame the selection of individuals to evaluate more dynamically, using an idea that can be found in noisy optimisation algorithms, where true fitness values are not known either. Of course, in case of noisy fitness functions, there is no option to evaluate the individual exactly, which calls for different algorithms. However, the partial order defined in \cite{Rudolph2001} is based on intervals for noisy single-objective fitness values and bears some semblance to one variation of the SAPEO algorithm detailed in section \ref{sec:formal}.

\subsection{Benchmarking}
\label{sec:bbob}

\begin{figure}[htb!]
    \centering
    \includegraphics[width=0.6\textwidth]{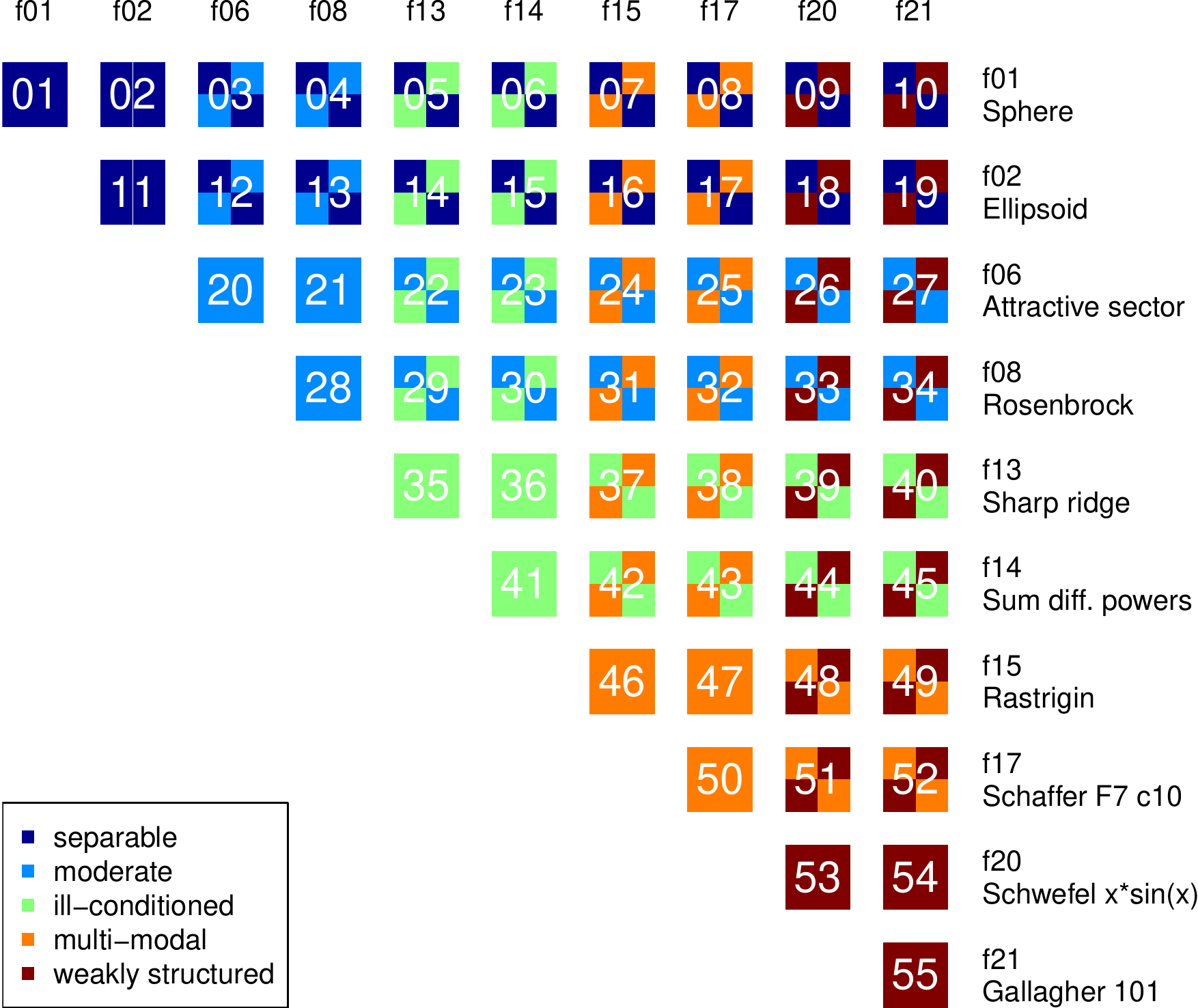}
    \caption{The 55 BBOB-BIOBJ functions are combinations of 10 single-objective functions (on the top and right). The groups the single-objective and the resulting bi-objective functions belong to are colour-coded according to the legend.}
    \label{fig:figures}
\end{figure}

BBOB-BIOBJ is a bi-objective Black-Box Optimisation Benchmarking test suite \cite{Tusar2016}. It consists of 55 bi-objective functions that are a combination of 10 of the 24 single-objective functions in the BBOB test suite established in 2009 \cite{Hansen2009}. In order to obtain a diverse test suite and to benchmark general algorithm performance across function types, the single-objective functions were chosen to maximise their diversity considering separability, conditioning, modality and global structure \cite{Hansen2009}. Based on these properties, the single-objective functions are divided into 5 function groups, from which 2 functions are chosen each. The resulting problems and corresponding properties are visualised in figure \ref{fig:figures}.

In an effort to measure performance on the different function types more accurately, each of the functions in the bi-objective test suite has 10 instances that, among other properties, differ in terms of the location of optima. As an effect, the scale of the optima and achievable absolute improvement of the objective values also vary significantly across instances (thus also between objectives). An algorithm with robust performance for all instances can therefore be evaluated with a higher confidence when not tested on a sole member of a function group.

All of the functions in the test suites are defined for search spaces of multiple dimensions, of which we will be considering dimensions $2,3,5,10$ and $20$ in order to be able to evaluate a wide range of problem sizes. The search space of each function is limited to $[-100,100] \subset \mathbb{R}$ per dimension in BBOB-BIOBJ.

The performance of an algorithm on the benchmarking suite is measured using a quality indicator expressing both the size of the obtained Pareto set and the proximity to a reference front. Since the true Pareto front is not known for the functions in the test suite, an approximation is used obtained by combining all known solutions from previous algorithmss. The ideal and nadir points are known, however, and used to normalise the quality indicator to enable comparisons across functions \cite{Brockhoff2016}. The metric reported as a performance measure for the algorithm is called precision and is the difference of the quality indicator of the reference set $I_{ref}$ and the indicator value of the obtained set. 58 target precisions are defined and the number of function evaluations needed to obtain the required approximation (or improvement) of the reference set is reported during a benchmark run. This way, the COCO platform enables an anytime comparison of algorithms, i.e. an evaluation of algorithm performance for each target precision and number of function evaluations \cite{Brockhoff2016}.

\section{Surrogate-Assisted Partial Order-Based Evolutionary Optimisation}

\subsection{Formal description}
\label{sec:formal}
Let $\tilde{f}(X_i) \in \mathbb{R}^d$ be the predicted fitness for individual $X_i$ with
\begin{align*}
	\tilde{f_k}(X_i)=f_k(X_i)+ e_{i},\, e_{i} \sim \mathcal{N}(0, \tilde{\sigma_i}), \, k \in \{ 1\dots d \}.
\end{align*}
Assuming that the uncertainty $\tilde{\sigma_i}$ was estimated correctly (\textbf{A1}), it follows that
\begin{align}
	&\mathbb{P}\Bigl( \,f_k(x_i) \in [\tilde{f_k}(x_i) - u_i, \tilde{f_k}(x_i) + u_i] \,\Bigr)= 1-\alpha \; \text{with} \nonumber \\
	&u_i = \tilde{\sigma_i} z \left(1-\frac{\alpha}{2}\right) \label{eq:u}
\end{align} since $P(|e_i|\leq u_i)= 1-\alpha$ and where $z$ denotes the quantile function of the standard normal distribution. In case the objective values are stochastically independent, which is true for the BBOB-BIOBJ benchmark, we can therefore conclude that $f(X_i)$ lies within the hypercube bounded by the described interval in each dimension with probability $(1-\alpha)^d$.

Assuming the function values lie within the defined hypercubes (\textbf{A2}), we can distinguish individuals confidently just based on the predicted hypercubes. Of course, in case of large uncertainties $\tilde{\sigma_i}$ and the resulting large hypervolume of the hypercubes, a distinction cannot be meaningful. SAPEO therefore introduces a threshold $\epsi_g$ for the uncertainty that is adapted in each iteration $g$ of the MOEA and decreased over the runtime of the algorithm depending on the distinctness of the population (more details in section \ref{base}). Individuals $X_i$ where $u_i > \epsi_g$ are evaluated exactly in generation $g$.

To distinguish between individuals, we propose binary relations incorporating the information on confidence bounds to varying degrees. All of these relations induce a strict partial order (irreflexivity, transitivity) on a population, including and akin to the Pareto dominance commonly used in MOEAs for the first part of the selection process. We analyse the proposed relations in terms of the probability and magnitude of a \textbf{sorting error $e_o$}. Our definition of a sorting error is always viewed in comparison to the partial order induced by the Pareto dominance. We define the probability  $\mathbb{P}(e_{o,r}^{i,j})$  of a sorting error made by relation $\preceq_r$ on individuals $X_i, X_j$ and the magnitude of the error $e_{o,r}^{i,j}$:
\begin{align*}
	\mathbb{P}(e_{o,r}^{i,j}) &= \mathbb{P} \Bigl( \, X_i \not\preceq X_j | X_i \preceq_r X_j \, \Bigr)\\
	e_{o,r}^{i,j} &= \begin{cases} | f(X_i)-f(X_j) | & \text{if } (X_i \preceq_r X_j) \wedge (X_i \not\preceq X_j)\\
	0 & \text{else} \end{cases} 
\end{align*}
A single or more sorting errors can but do not have to lead to \textbf{selection errors} $e_s$, where the individuals selected based on the dominance relation and a secondary criterion defined below differ from the ones selected with Pareto dominance and hypervolume as secondary criterion.

\paragraph{$\preceq_f$: Pareto dominance on function values} This relation is the standard in MOEAs and it is obvious that $\mathbb{P}(e_{o,f}^{i,j})=0 \; \forall i,j$.
\begin{align*}
X_i \preceq_f X_j := f(X_i) \preceq f(X_j)
\end{align*}
\paragraph{$\preceq_u$: Confidence interval dominance} Assuming \textbf{A2}, if
\begin{align*}
X_i \preceq_u X_j := \bigwedge_{k\in [1 \dots d]} \tilde{f_k}(X_i)+u_i < \tilde{f_k}(X_j)-u_j
\end{align*}
holds, it is guaranteed that $X_i \prec X_j$. Assuming stochastic independence of the errors on predicted uncertainty, we can compute an upper bound for the probability of sorting errors per dimension:
\begin{align*}
	\mathbb{P}(e_{o,u}^{i,j}(k)) & = 	\mathbb{P} \Bigl(\, f_k(X_i) \geq f_k(X_j) | X_i \preceq_u X_j \, \Bigr)\\
	& \leq \mathbb{P} \Bigl(\,f_k(X_i) \geq \tilde{f_k} (X_i) + u_i \vee f_k(X_j) \leq \tilde{f_k}(X_j)-u_j \, \Bigr)\\
	& \leq \frac{\alpha}{2} + \frac{\alpha}{2} = \alpha
\end{align*}
Only if the confidence hypercubes of two individuals intersect, the probability of them being incomparable is greater than 0. Because of the way $X_i \preceq_u X_j$ is defined, this is only possible if a sorting error is made in every dimension. It follows that $\mathbb{P}(e_{o,u}^{i,j}) \leq \alpha^d \, \forall i,j$ assuming \textbf{A1}, making sorting and, as a consequence, selection errors controllable.
\paragraph{$\preceq_c$: Confidence interval bounds as objectives} Another way of limiting the prediction errors potentially perpetuated through the algorithm is to limit the magnitude of the sorting error. For this reason we define:
\begin{align*}
X_i \preceq_c X_j := &
	\bigwedge_{k\in [1 \dots d]} \left(\begin{smallmatrix}
\tilde{f_k}(X_i)-u_i \\ \tilde{f_k}(X_i)+u_i
\end{smallmatrix} \right) \precsim
\left(\begin{smallmatrix}
\tilde{f_k}(X_j)-u_j \\ \tilde{f_k}(X_j)+u_j
\end{smallmatrix} \right)\\
&\wedge \exists k \in [1\dots d]: \left(\begin{smallmatrix}
\tilde{f_k}(X_i)-u_i \\ \tilde{f_k}(X_i)+u_i
\end{smallmatrix} \right) \preceq
\left(\begin{smallmatrix}
\tilde{f_k}(X_j)-u_j \\ \tilde{f_k}(X_j)+u_j
\end{smallmatrix} \right)
\end{align*}
Under assumption \textbf{A2}, the error per dimension is bounded by the length of intersection of the confidence intervals, which is in turn bounded by the width of the smaller interval. Therefore, it holds that $e_{o,c}^{i,j}\leq 2^d \min(u_i,u_j)^n \, \forall i,j$.
\paragraph{$\preceq_p$: Pareto dominance on predicted values} This relation is the most straightforward, but it does not take the uncertainty of the prediction into account.
\begin{align*}
X_i \preceq_p X_j := \tilde{f}(X_i) \preceq \tilde{f}(X_j)
\end{align*}
Assuming \textbf{A2} again, a sorting error can only be committed if the confidence intervals intersect. Because of the symmetric nature of the interval $e_{o,p}^{i,j} \leq (u_i+u_j)^d \, \forall i,j$ holds, as the magnitude of the sorting error is again bounded by the confidence interval widths.
\paragraph{$\preceq_o$: Pareto dominance on lower bounds} This optimistic relation was motivated by empirical findings that the performance of SA-MOEAs can be improved using $\preceq_o$ instead of $\preceq_p$.
\begin{align*}
X_i \preceq_o X_j := & \bigwedge_{k\in [1 \dots d]} \tilde{f_k}(X_i)-u_i 
 \leq
\tilde{f_k}(X_j)-u_j\\
& \wedge \exists k \in [1 \dots d]: \tilde{f_k}(X_i)-u_i < \tilde{f_k}(X_j)- u_j
\end{align*}
The maximum error occurs when the lower confidence interval bounds are close together, but in the wrong order, making $e_{o,o}^{i,j} \leq 2^n \max(u_i,u_j)^d \, \forall i,j$.

Now assume we have obtained a strict partial order based on any of the given binary relations. Let $r_c$ be the rank of the $\mu$-th individual. Then, all individuals with rank less than $r_c$ can confidently (with maximum errors as described above) be selected. In case a selection has to be made from the individuals with the critical rank $r_c$, one option is to apply another dominance relation to the individuals in question in hopes that the required distinction can be made.

Another option is to use a relation inducing a total preorder (transitivity, totality) as a secondary selection criterion (and random choice in case of further ties) as most MOEAs do. Again incorporating different information, we have tested the following hypervolume-based (\textit{hv}) relations for this purpose:
\begin{compactitem}
	\item Hypervolume contribution of objective values:\\
	$X_i \leq_{ho} X_j := hv (f_o(X_i)) \geq hv (f_o(X_j))$, where $f_o(X_i) = \begin{cases}
		f(X_i) & u_i =0\\
		\tilde{f}(X_i) & \text{else}
	\end{cases}$
	\item Hypervolume contribution of confidence interval bounds:\\ $X_i \leq_{hc} X_j :=  \prod_{k\in [1 \dots d]} hv \bigl(\begin{smallmatrix}
\tilde{f_k}(X_i)-u_i \\ \tilde{f_k}(X_i)+u_i
\end{smallmatrix} \bigr) \geq
\prod_{k\in [1 \dots d]} hv \bigl(\begin{smallmatrix}
\tilde{f_k}(X_j)-u_j \\ \tilde{f_k}(X_j)+u_j
\end{smallmatrix} \bigr)$
\end{compactitem}

\subsection{SAPEO algorithm}
\label{base}

\begin{algorithm}[htb] 
  \caption{SAPEO} 
  \label{alg1} 
  \begin{algorithmic}[1] 
    \REQUIRE fun, local\_size, budget, strategies, scnd\_crit  
    \ENSURE $X_{final}$ \COMMENT{final population}
    \STATE $X_0.genome \Leftarrow [random(n): i \in 1 \dots pop\_size]$ \COMMENT{Random initialisation} \label{initX}
    \STATE $X_0.f \Leftarrow [fun(x): x \in X_0]; \, X_0.e =0$  \COMMENT{evaluate sampled
      individuals}  \label{initF}
    \STATE $O \Leftarrow init(X_0,budget)$ \COMMENT{initialise optimiser
      with initial population}  \label{initO}
    \STATE $\epsi_0 \Leftarrow \infty; \, g \Leftarrow 1$ \COMMENT{initialise error
      tolerance and generation counter} \label{initEpsi}
    \WHILE{(\NOT $O.stop())$ \OR ($\epsi > 0 \, \AND \, budget>0$)} \label{loopStart}
       \STATE $X_g.phenome \Leftarrow O.evolve(X_{g-1})$ \COMMENT{get new population} \label{newPop}
       \STATE $X_g.f, X_g.e \Leftarrow [model(x, knearest(x, X[X.e==0],local\_size)): x \in X_g]$ \COMMENT{predict value and error based on local surrogate from evaluated neighbours} \label{predictF}
       \STATE $\epsi_g \Leftarrow \min(e_{g-1}, \alpha\text{-percentiles}(\text{diff}(X_g.f))$ \COMMENT{update
         error tolerance} \label{updateEpsi}
       \FORALL{$x \in X_g$} 
       		\IF{$x.e>\epsi_g \, \OR \, O.select(X_g,strategy,scnd\_crit) == \text{NULL}$}\label{REV}
				\STATE $x.f = fun(x)$ \COMMENT{evaluate individual} \label{updateF}
				\STATE $bbob.recommend(X[X.e>0, \text{last}])$  \COMMENT{recommend solution} \label{rec}
       		\ENDIF
       \ENDFOR \label{endREV}
       \STATE $X_{g} = O.select(X_g,strategy,scnd\_crit)$ \COMMENT{SAPEO survival
         selection} \label{selectO}
       \STATE $g=g+1$ \COMMENT{increase generation counter}
    \ENDWHILE
  \STATE $X_{final} = X_{g-1}; X_{final}.f = [fun(x): x \in X_{final}]$ \COMMENT{evaluate final population} \label{finalEval}
  \end{algorithmic}
\end{algorithm}

\clearpage
Algorithm \ref{alg1} describes the basic SAPEO algorithm, which any MOEA and surrogate model with uncertainty estimates can be plugged into. As inputs, the algorithm receives the fitness function \lstinline|fun|,
the number of points considered for the surrogate model \lstinline|local_size|, the \lstinline|budget| of function evaluations as well as the order of dominance relations \lstinline|strategies| and the secondary criterion \lstinline|scnd_crit| to be used for selection as described above. The output is the final population.

The algorithm starts with the initialisation of mandatory data structures; the population is initialised randomly (line
  \ref{initX}), these points are evaluated using the considered fitness function
  \lstinline|fun| (line \ref{initF}), the MOEA is initialised (line \ref{initO}) and the error tolerance $\epsi$ and generation counter are initialised (line \ref{initEpsi}).

The core optimisation loop starts in line \ref{loopStart} and stops if either the considered optimiser terminates (due to the allocated budget or convergence), but not while the error tolerance $\epsi$ is larger than $0$ and there are function evaluations left. 
Within this loop, new candidate solutions $X$ are first generated by the optimisation algorithm (line \ref{newPop})  and evaluated based on a local surrogate model trained from the $local\_size$ evaluated individuals closest in design space (line \ref{predictF}). The predicted function values are stored as well as the expected model errors computed using equation \ref{eq:u}. The error tolerance threshold is then adapted (line \ref{updateEpsi}) based on the euclidian distances in objective space between all individuals in the population per dimension, since the goal is to be able to distinguish solutions without exact evaluations. We reduce the threshold during the course of the algorithm in order to limit the risks of sorting errors with large effects on the final population. Therefore, $\epsi_g$ is the minimum of the previous threshold $\epsi_{g-1}$ and the $\alpha$-percentiles of the euclidian distances in objective space per dimension.

If any of the individuals in the population need to be evaluated - either because the predicted uncertainty is above the threshold or because the individuals can not be distinguished otherwise (see line \ref{REV}) - they are evaluated in line \ref{updateF} and the individuals' fitness values are updated accordingly. In order to simulate anytime behaviour of the algorithm, each time a solution is evaluated, the last unevaluated individual is recommended to the BBOB framework in line \ref{rec} (i.e. the current number of function evaluations is recorded). The reasoning is that, since the goal of the algorithm is to delay the evaluation of individuals as much as possible, it would be unfair not to consider the individuals at the time they are found. If the algorithm stops after function evaluation $i$, the current population would still be evaluated in the following evaluations $i+1 \dots i+$population size. This behaviour is approximated here by recording the most promising solutions at the appropriate number of function evaluations.

The set of candidate solutions, along with the (predicted but reasonably certain) function values and the expected prediction errors are then passed to the optimiser in line \ref{selectO}. Depending on the selected strategy, the optimiser then selects the succeeding population as described above with regard to the predicted function values and uncertainties and resumes its regular process.

Finally, after the optimisation loop terminates, the function values of the individuals in the final population are computed using the real fitness function in line \ref{finalEval}, in case there are any individuals left that have not been evaluated.

\section{Evaluation}

\subsection{Experimental Setup}
The maximum budget was set to $1000 \times n$ in order to keep with the theme of expensive functions. Since the performance is strictly measured in terms of function evaluations (target precision reached per function evaluation, cf. section \ref{sec:bbob}), the runtime does not influence it. We did run each experiment with 550 parallel jobs that took less than 3 hours each.

We compare the performances with a standard \cite{Beume2007} and surrogate-assisted SMS-EMOA with pre-selection as proposed by \cite{Emmerich}, since we are not aware of any SA-MOEAS using the SMS-EMOA with individual-based surrogate management strategies. Specifically, we look at the following algorithms:
\begin{compactitem}
	\item[\textbf{SMS-EMOA}] Standard SMS-EMOA as baseline comparison.
	\item[\textbf{SA-SMS-p}] Surrogate assisted SMS-EMOA using $\preceq_p$ for pre-selection.
	\item[\textbf{SA-SMS-o}] Surrogate assisted SMS-EMOA using $\preceq_o$ instead (experimentally shown to improve the performance of pre-selection for the NSGA-II \cite{Emmerich}).
	\item[\textbf{SAPEO-uf-ho}] SAPEO using $\preceq_u$ to rank the offspring, thus accepting a risk of sorting errors of only $\alpha^2$ (cf. \ref{sec:formal}). For as long as the population can not be distinguished by $\preceq_u$, the invididuals are evaluated according to $\preceq_f$, thus avoiding making any further sorting errors. The hypervolume relation $\leq_{ho}$ is used as secondary criterion. This algorithm should therefore only take small risks and behave like the SMS-EMOA while saving function evaluations.
	\item[\textbf{SAPEO-ucp-ho}] SAPEO using increasingly risky relations $\preceq_u,\, \preceq_c, \, \preceq_p$ to avoid evaluations completely if not forced by the uncertainty threshold $\epsi$, taking the opposite approach as SAPEO-uf-ho. $\leq_{ho}$ is used as secondary criterion.
	\item[\textbf{SAPEO-uc-hc}] SAPEO using multi-objectification of the confidence interval\\boundaries fully. It uses $\preceq_u$ as a first safer way of ranking, followed by $\preceq_c$ on critical individuals. Secondary criterion is $\leq_{hc}$.
\end{compactitem}

In order to enable a direct comparison, we use the same parametrisations and operators wherever possible across all algorithms. We use the standard variation operators and corresponding parametrisation \cite{Beume2007} and a population size of $100$ as suggested in \cite{Emmerich}. Due to practical concerns about computation time, we limit the sample size for the local surrogate models to $15$. In a real-world application, the sample size should be chosen considering the tradeoff between computation times for the model and the fitness function. The number of candidate offspring for the SA-SMS algorithms was set to $15$ as well. The Kriging models assume a squared exponential correlation and a constant trend. The regression weight vectors are estimated with the maximum likelihood computed by the COBYLA algorithm according to universal Kriging starting from $10^{-2}$ with $10^{-4}$ as lower and $10^{1}$ as upper bounds.

\clearpage

\subsection{Visualisation and Interpretation of the Results}

There are two main angles to evaluating the anytime performance of algorithms - either measuring the performance indicator at predefined numbers of function evaluations (\textit{fixed budget}) or recording at which function evaluation target performances have been reached (\textit{fixed target}). In the following, we use the latter as the resulting measures are quantifiable.	

\begin{figure}[htb]
    \centering
    \includegraphics[width=0.8\textwidth]{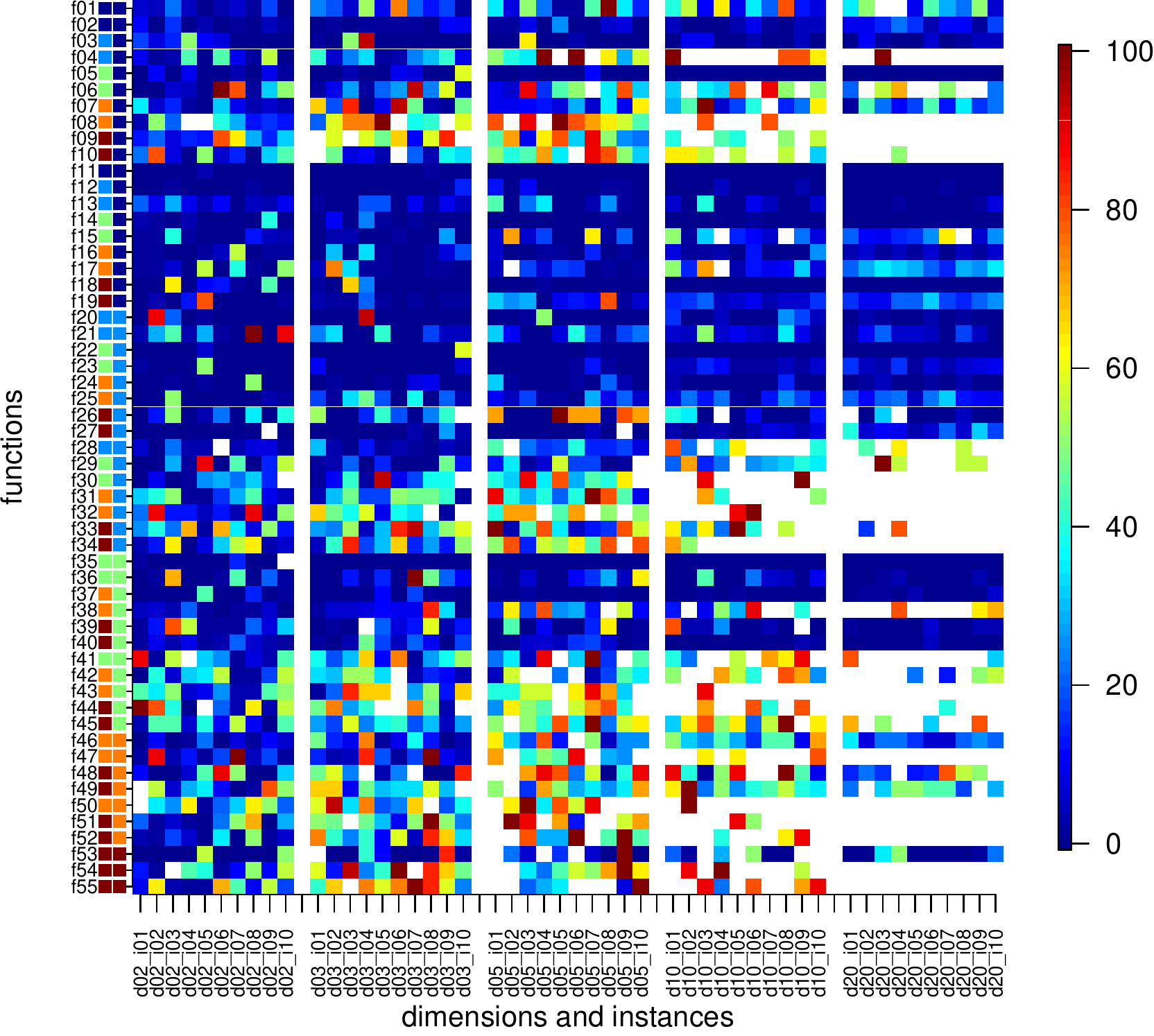}
    \caption{SAPEO-uf-ho performance in terms of the percentage of the budget per dimension used to reach target $1$ for all function instances and dimensions}
    \label{fig:sapeobig}
\end{figure}

For a detailed depiction of an algorithm's performance for a fixed target, we use heatmaps like figure \ref{fig:sapeobig} that show the percentage of budget per dimension used until the target was reached according to the colour scale on the right. The dimensions and instances of each function are shown separately to be able to analyse how well the algorithm performance genaralises across function instances and dimensions. This is very important to justify the aggregation of performance measures across instances. The functions have colour codes according to the legend in figure \ref{fig:figures} that specify the function groups they belong to. If the target is never reached within the specified budget, the corresponding square is white.

From the plot, it is apparent that for the selected target $1$, the algorithm SAPEO-uf-ho has trouble with a number of functions even in small dimensions. Additionally, for those functions, the algorithm's performance seems to drop with increasing dimension of the search space. Especially the Rosenbrock function seems to be problematic for the algorithm: SAPEO-uf-ho rarely reaches the target for dimensions 10 or 20 when the Rosenbrock function is part of the problem (f 04, 13, 21, 28-34). A potential explanation could be that the surrogate model is not able to represent the narrow valley containing the optimum accurately enough. Another explanation could be that the underlying SMS-EMOA just does not deal well with the Rosenbrock function, possibly because of the variation operators used. As expected, some of the weakly structured problems were difficult for SAPEO-uf-ho as well, likely again due to the surrogate model.

However, in order to be able to discuss the potential causes of poor performances, the comparison to the other algorithms has to be considered. In order to get a better overview of the performances of all algorithms and to be able to detect patterns, we have compiled figure \ref{fig:mega}, which is an assembly of 30 heatmaps like the one in figure \ref{fig:sapeobig} for all algorithms and different targets. The same colour scale as in figure \ref{fig:sapeobig} is used, recall that white spaces signify targets that were not reached within the allocated budget.
\input{megaFigure}

In figure \ref{fig:mega}, the most obvious trend is the declining performance for each target precision, which was of course expected. It is also quite apparent that the SMS-EMOA performs better in general than all other algorithms for each target precision. However, we can also see that all SAPEO versions are an improvement when compared to the SA-SMS algorithms. Interestingly, we also see similar patterns in terms of which functions are more difficult for all algorithms.

\begin{figure}[htb!]
	\centering
	\includegraphics[width=1\textwidth]{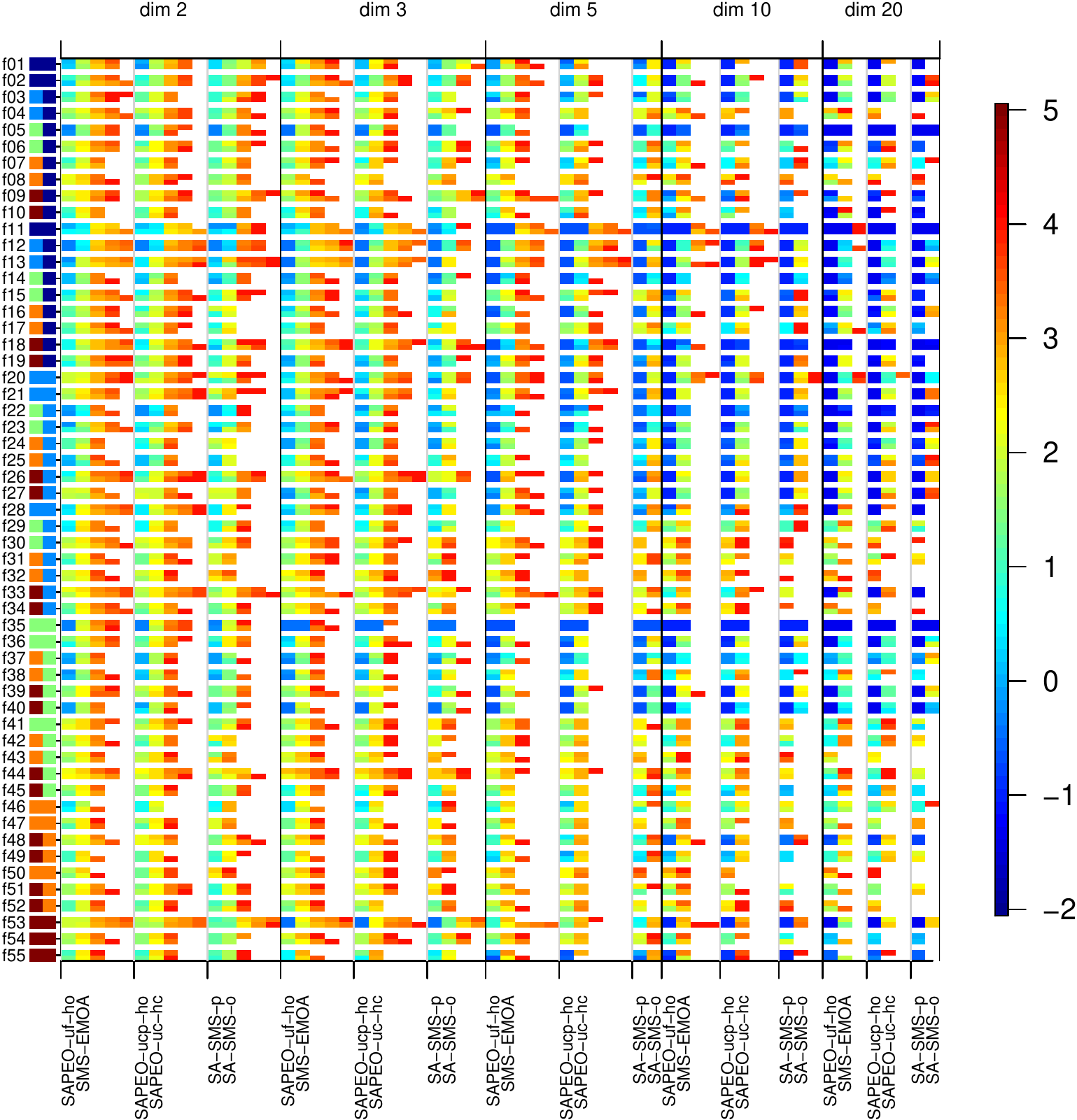}
	\caption{BBOB-BIOBJ performance results for all algorithms regarding expected runtime (colour coded in log-scale) for targets $10^1, 10^0, 10^{-1}, 10^{-2}, 10^{-3}$}
	\label{fig:aggregated}
\end{figure}

Unfortunately, while providing a good overview, figure \ref{fig:mega} is not well suited to interpret the performance of each algorithm per function. While very detailed, the plots are not easy to interpret due to the abundance of information displayed at once. In order to analyse the circumstances of different performance patterns, we compile a plot that aggregates over the different instances of a function. This way, the general performance of an algorithm per function can be expressed without risk of overfitting, as intended by the COCO framework. To do that, we use the \textit{expected runtime} (expected number of function evaluations) to reach a target  \cite{Hansen2010} as a performance measure. The measure is estimated for a restart algorithm with 1000 samples. The results are again displayed in a heatmap (figure \ref{fig:aggregated}). The colour visualises the estimated expected runtime per dimension according to the scale on the right in log-scale. Higher values than the maximum budget come to pass if a target is not reached in all instances. White spaces occur if the target was never reached by the algorithm in all instances.

The plot displays expected runtime for all dimensions in different columns according to the labels above. Each of these columns is again divided into 3, displaying the results for different algorithms according to the labels on the bottom. There are two algorithms per column, whose results are displayed on top of each other for each row corresponding to a function. For each algorithm, the expected runtimes for targets $10^1, 10^0, 10^{-1}, 10^{-2}, 10^{-3}$ are depicted in that order. In case a target was never reached for all algorithms in a column, it is omitted. For example, the expected runtime for SAPEO-uf-ho on function 01, target $10^1$ and dimension 2 is on the top left corner and encoded in a light blue. Therefore, the expected runtime to reach target $10^1$ is around $10^{0}*2=2$. The SMS-EMOA is directly below that and a shade lighter, so has a slightly higher expected runtime. Like in figure \ref{fig:sapeobig}, the groups each function belongs to are encoded according to the colour scheme in the legend of figure \ref{fig:figures}.

The general trends as seen in figure \ref{fig:mega} can be observed here as well. However, we can also see that SAPEO-uf-ho beats the SMS-EMOA in terms of precision reached on very rare occasions, for example on functions f03 and f41 in dimension 2 and function f20 in dimension 10. Still, the SMS-EMOA generally reaches the same or more precision targets than the other algorithms. However, in most cases where a higher precision target is reached by only a single algorithm, the corresponding colour indicates a very high expected runtime. This means that the algorithm didn't reach the higher target for most instances, which speaks against a robust performance on that algorithm. More importantly, the SA-SMS variants often reach less targets than the other algorithms, especially in higher dimensional problems, meaning they are clearly outperformed.

The colour gradients in most functions are remarkably alike, indicating similar behaviour and difficulties experienced with each problem. This is expected, as the intention of SA-MOEAs is to avoid function evaluations without influencing the evolutionary path too much. Possibly due to the aggregating nature of the expected runtime measure, the performance contrast does not appear to be as stark as in figure \ref{fig:mega}. The gradient and number of performance targets reached per function is in fact relatively similar for all algorithms. In most cases, differences occur towards the end of the gradient, indicating that the precision improvement of the surrogate-assisted algorithms is less steep than for the SMS-EMOA. However, in order to analyse the algorithms' behaviour appropriately in that regard, a more thorough analysis of the separate selection steps is required.

There seems to be no clear pattern on which algorithm works well for which kind of function from the plot. The different SAPEO versions vary only rarely. The performance of all algorithms seems to be more closely tied to the specific single-objective functions. For example, Rosenbrock seems to pose problems whereas Schwefel seems to be more manageable.



\subsection{Insights and Future Work}

Although the SAPEO variants perform better than both SA-SMS variants, the SMS-EMOA still beats them in most cases in terms of anytime performance. This fact is quite surprising, since especially the SAPEO-uf-ho variant should only take very small risks and make only minimally different decisions. A more thorough analysis is needed to ascertain the strengths and weaknesses of the algorithms, possibly using the function properties to train a machine learning model to predict controller performance and thus learning any observable patterns in performances.

One potential source of error is the surrogate model. First of all, the assumptions \textbf{A1, A2} (section \ref{sec:formal}) could be wrong. A large error in the predicted uncertainty could have a tremendous influence on the algorithm. Additionally, the uncertainties during the start of the SAPEOs were relatively large, which could also send the algorithm into a wrong direction. The uncertainties could be mitigated by using surrogate ensembles instead, distributing the samples better (e.g. with a latin hypercube), increasing the sample size or selecting a fitting kernel. There might also be better ways to adapt the uncertainty threshold $\epsi$. Additionally, the performances of local vs. global surrogates should be analysed more thoroughly. The fact that the SA-SMS approach performed even worse supports that possible explanation. These findings suggest that a major concern when using SA-MOEAs is the quality of the surrogate model, even when ignoring the computational costs.

Apart from the model, there are possible improvements regarding the binary relations used. For one, $\preceq_u$ could be defined without forcing strict Pareto dominance of the hypercubes. Furthermore, using hypercubes for the potential location of the fitness values is a simplification. Perhaps a binary relation on hyperellipsoids could provide better results.

Lastly, while the SAPEO approach worked well for single-objective problems, the same multi-objective problems pose an incomparably larger difficulty for a surrogate model. Besides, the algorithm used in the preliminary experiments was the CMA-ES, which uses $\mu+\lambda$ selection instead of $\mu +1$, as well as different variation operators. Therefore, it suggests itself to do a thorough study on SAPEO with single-objective problems first, analysing strength and weaknesses, before returning to the multi-objective case for further improvements.

%


\section{Conclusions}

In this paper, we have proposed a novel approach to surrogate-assisted multi-objective evolutionary algorithm called SAPEO. An extensive analysis of its anytime performance using the BBOB-BIOBJ benchmark showed that it was outperformed by its underlying algorithm in this case, the SMS-EMOA. However, SAPEO still beats the most comparable SA-MOEA on the benchmark. We suggest some improvements and future work in the previous section and conclude that the influence of the quality of surrogate models on SA-MOEAs should be analysed more carefully. With the proper techniques to learn a surrogate and corresponding parametrisation, SAPEO should be able to beat its underlying algorithm as it does on single-objective problems.

%% file: megaFigure.tex
\captionsetup[subfigure]{labelformat=empty, font=scriptsize}

\begin{figure}[b!]
  \begin{adjustbox}{angle=90}
  	\begin{subfigure}[b]{0.16\textwidth}
  		\hspace{2pt}
    	\caption{SAPEO-uc-hc}
  	\end{subfigure}
  \end{adjustbox}
  \begin{subfigure}[b]{0.18\textwidth}
    \caption{$10^1$}
    \includegraphics[width=\textwidth]{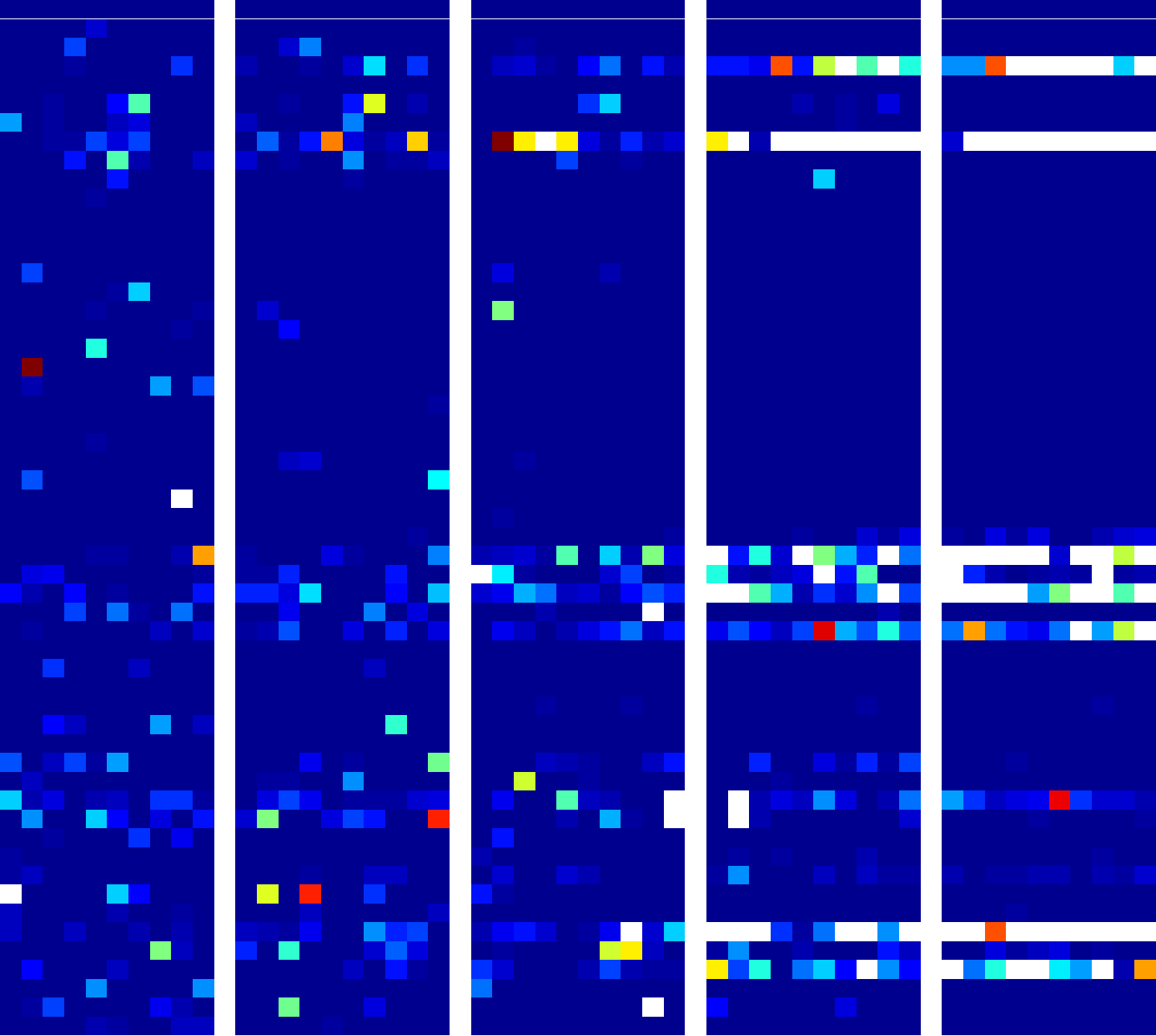}
  \end{subfigure}
  \begin{subfigure}[b]{0.18\textwidth}
    \caption{$10^0$}
    \includegraphics[width=\textwidth]{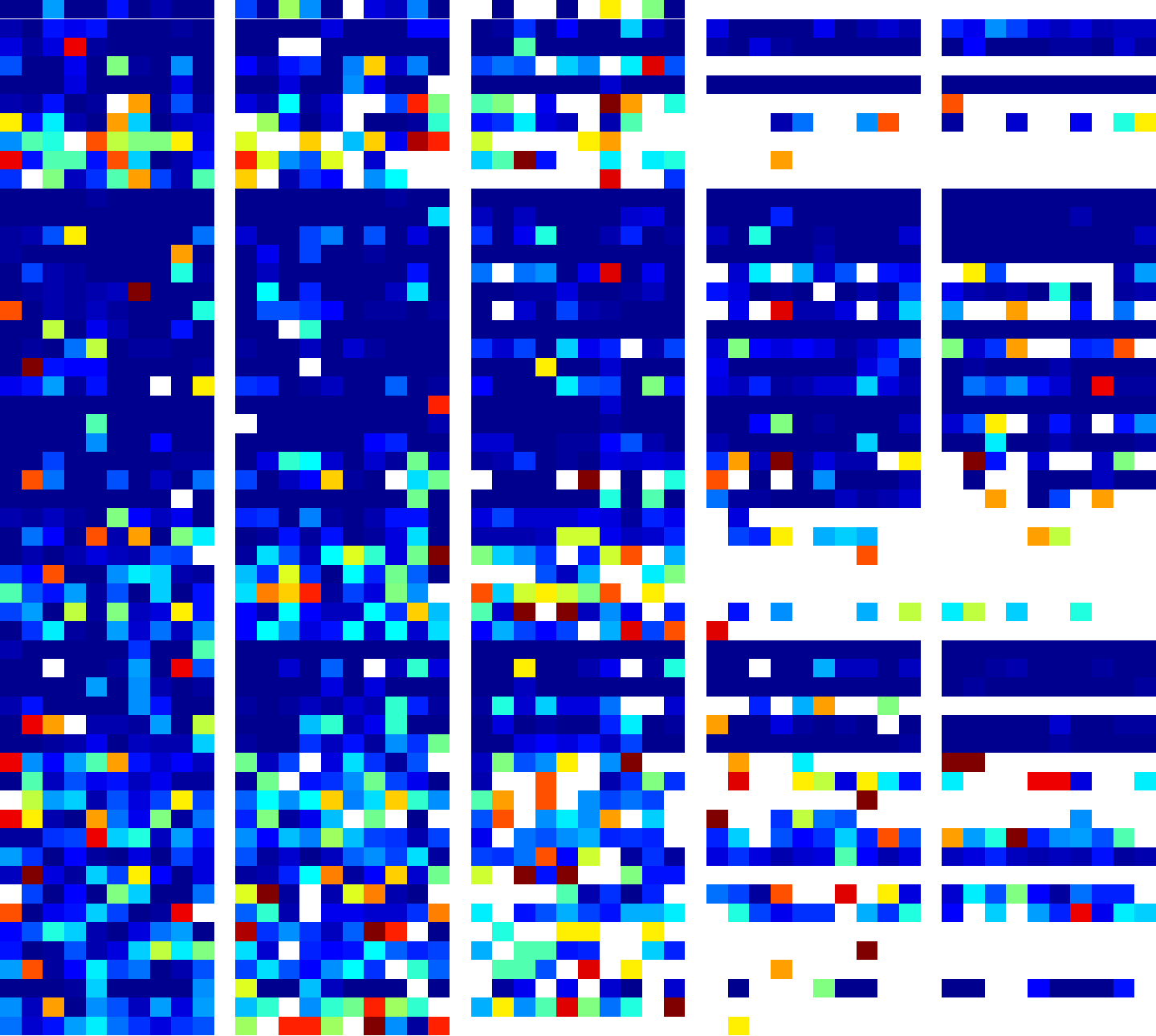}
  \end{subfigure}
    \begin{subfigure}[b]{0.18\textwidth}
    \caption{$10^{-1}$}
    \includegraphics[width=\textwidth]{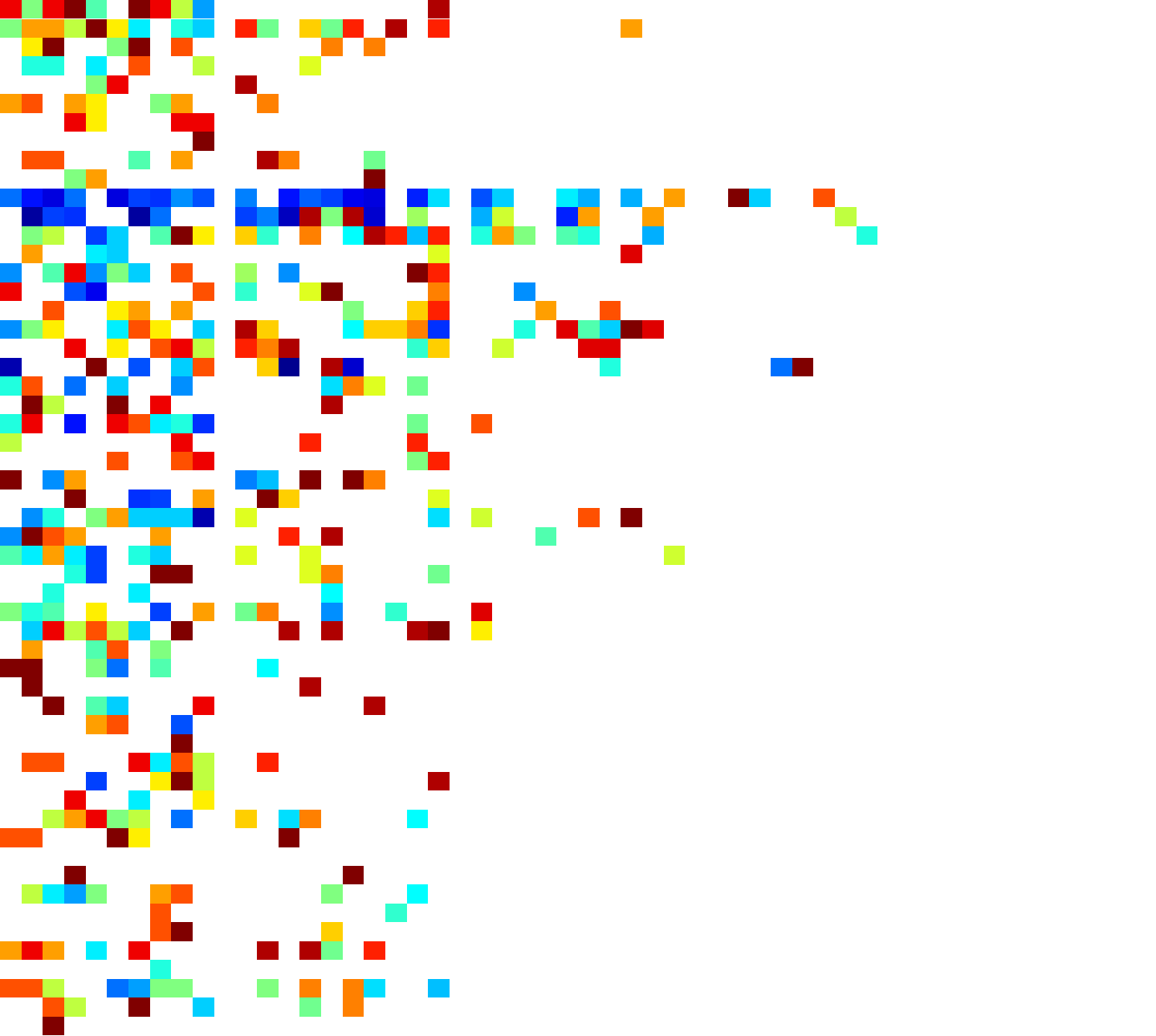}
  \end{subfigure}
    \begin{subfigure}[b]{0.18\textwidth}
    \caption{$10^{-2}$}
    \includegraphics[width=\textwidth]{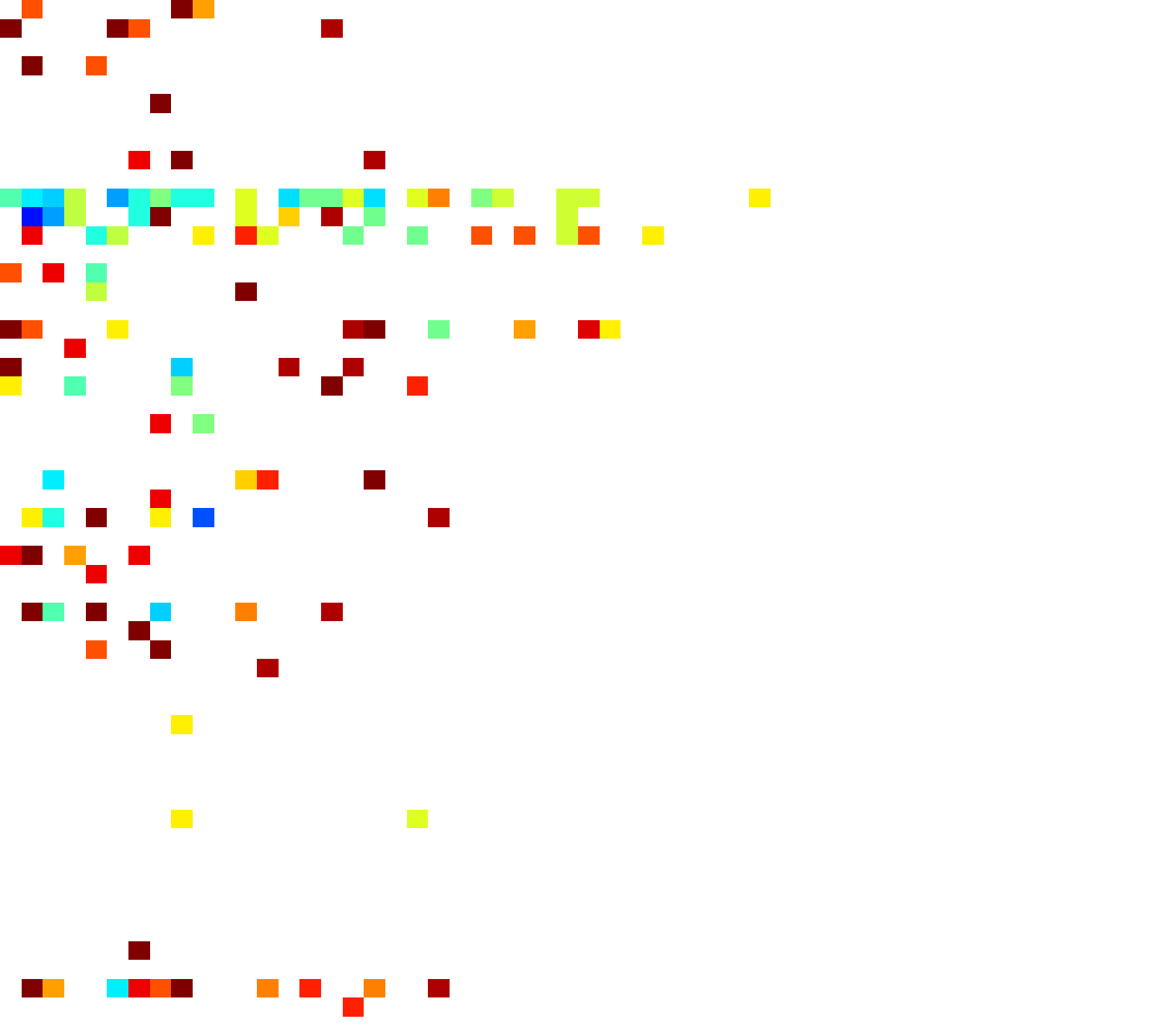}
  \end{subfigure}
    \begin{subfigure}[b]{0.18\textwidth}
    \caption{$10^{-3}$}
    \includegraphics[width=\textwidth]{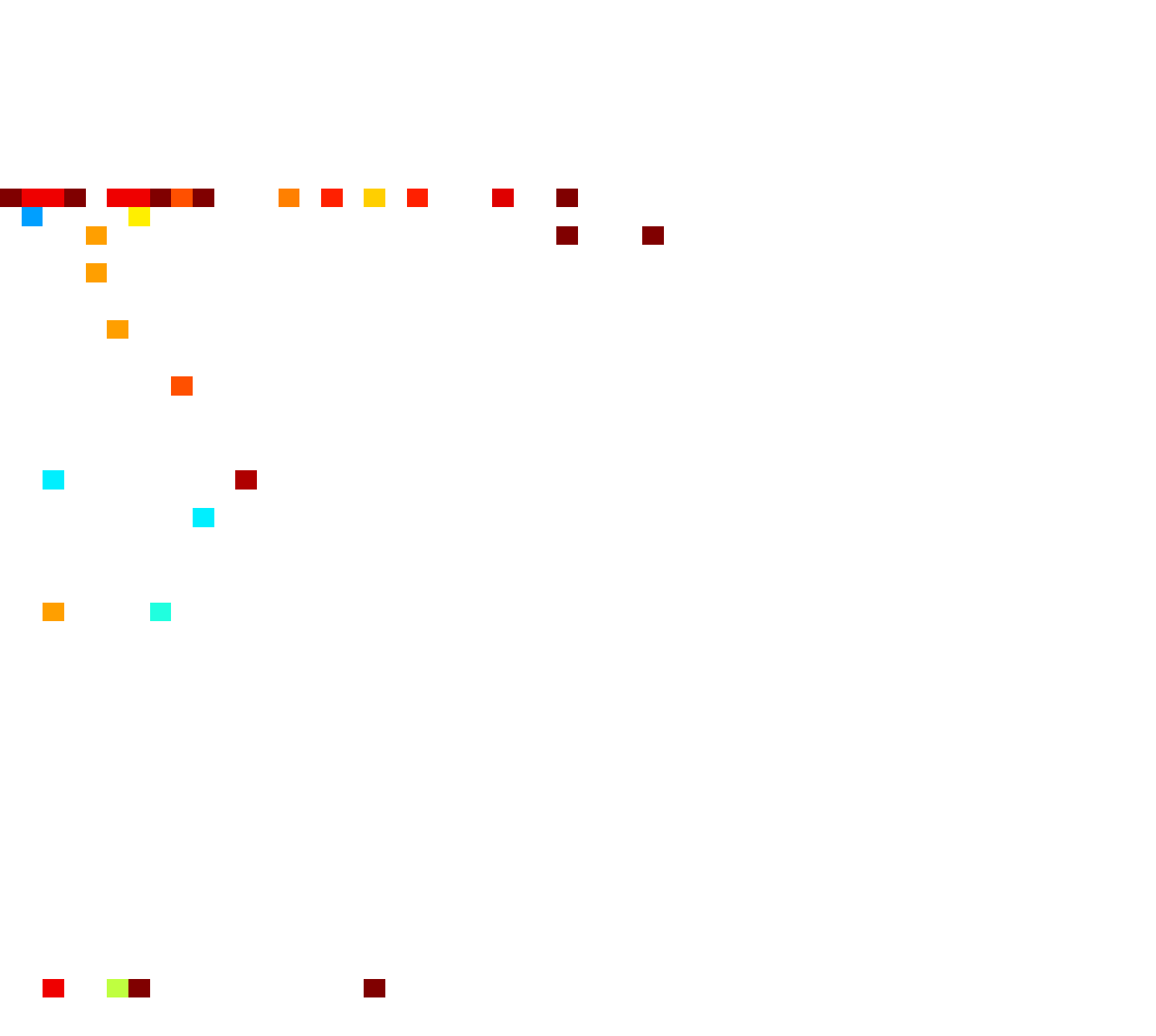}
  \end{subfigure}

  \begin{adjustbox}{angle=90}
  	\begin{subfigure}[b]{0.16\textwidth}
  		\hspace{2pt}
    	\caption{SAPEO-ucp-ho}
  	\end{subfigure}
  \end{adjustbox}
  \begin{subfigure}[b]{0.18\textwidth}
    \includegraphics[width=\textwidth]{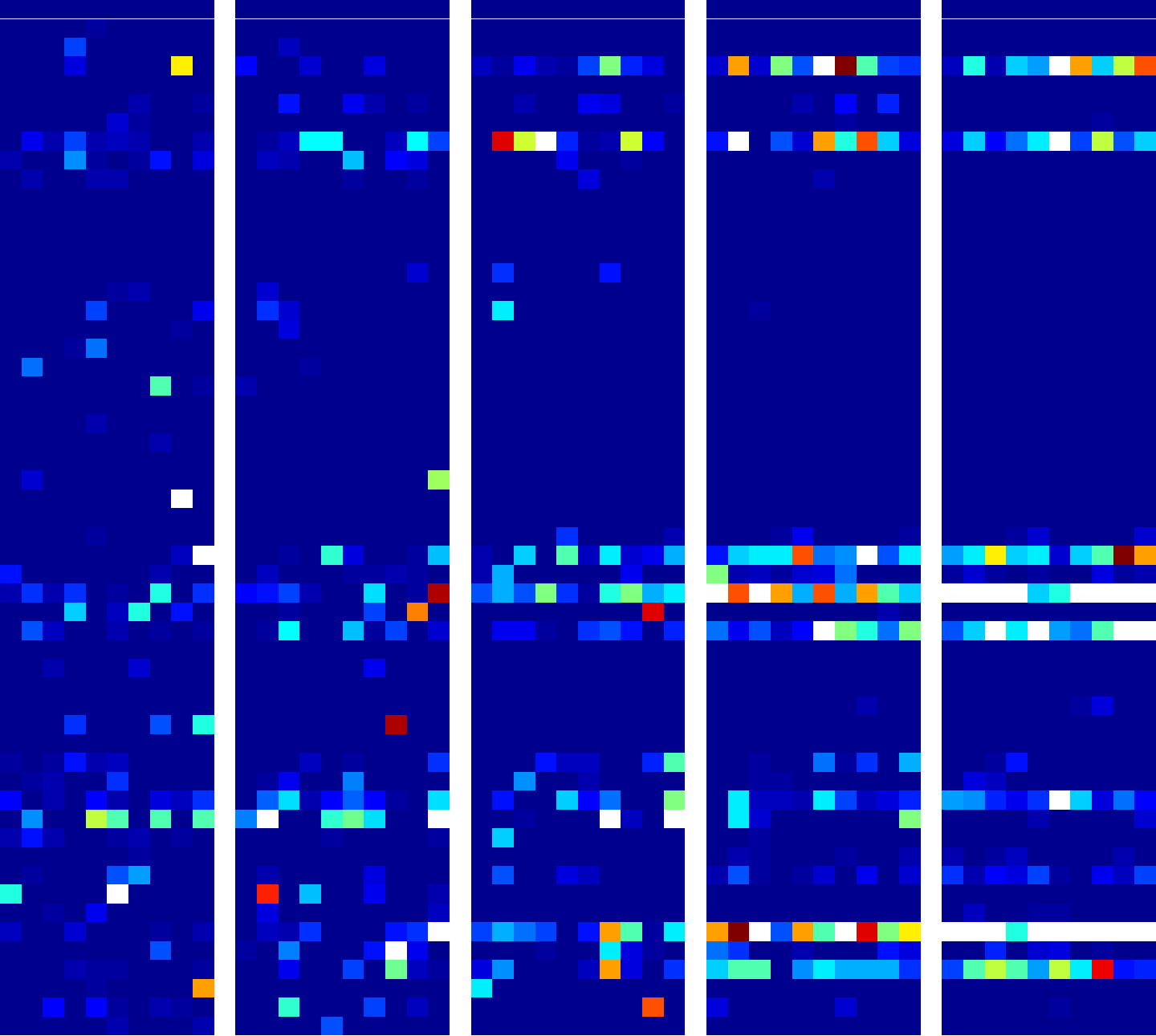}
  \end{subfigure}
  \begin{subfigure}[b]{0.18\textwidth}
    \includegraphics[width=\textwidth]{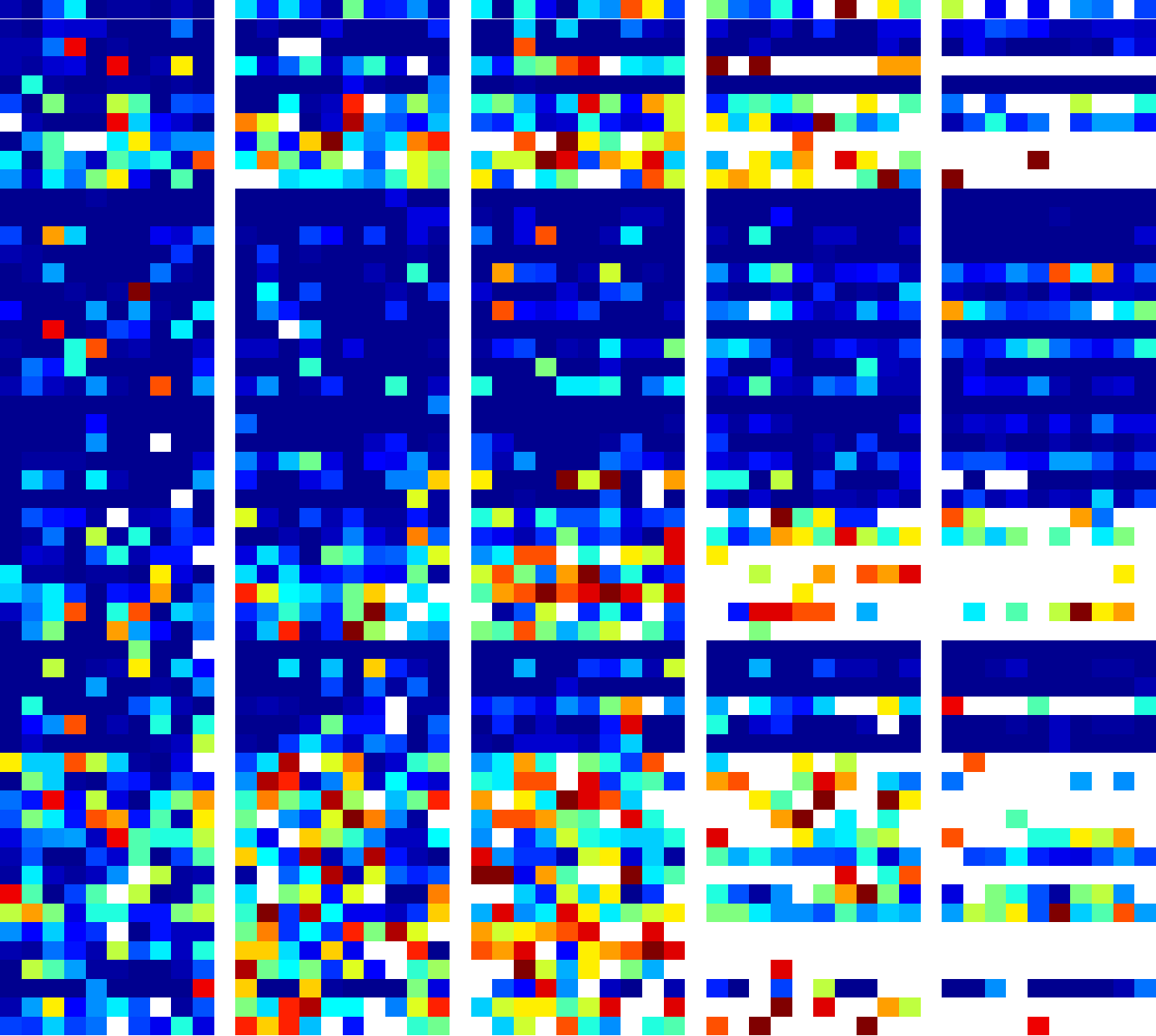}
  \end{subfigure}
    \begin{subfigure}[b]{0.18\textwidth}
    \includegraphics[width=\textwidth]{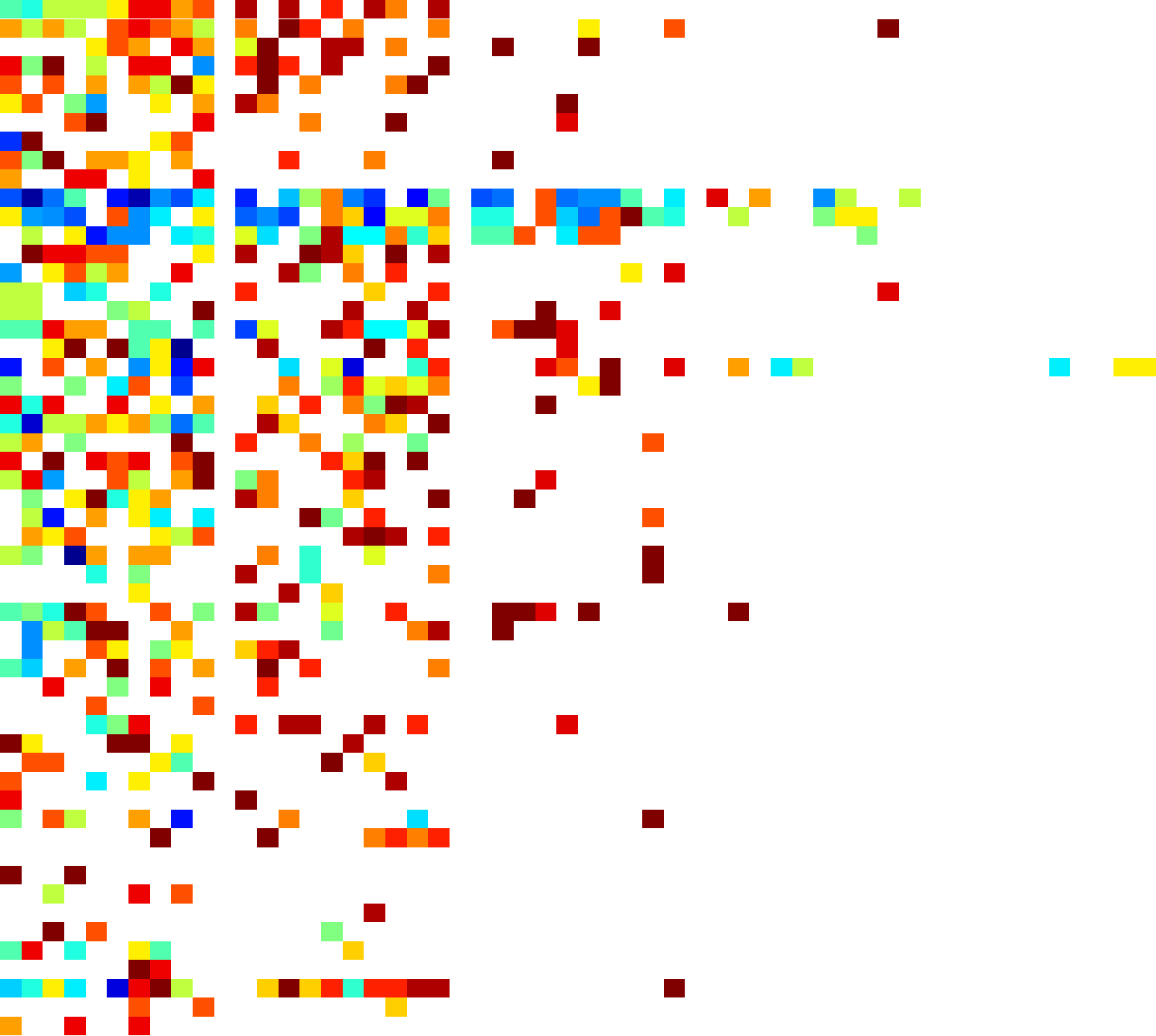}
  \end{subfigure}
    \begin{subfigure}[b]{0.18\textwidth}
    \includegraphics[width=\textwidth]{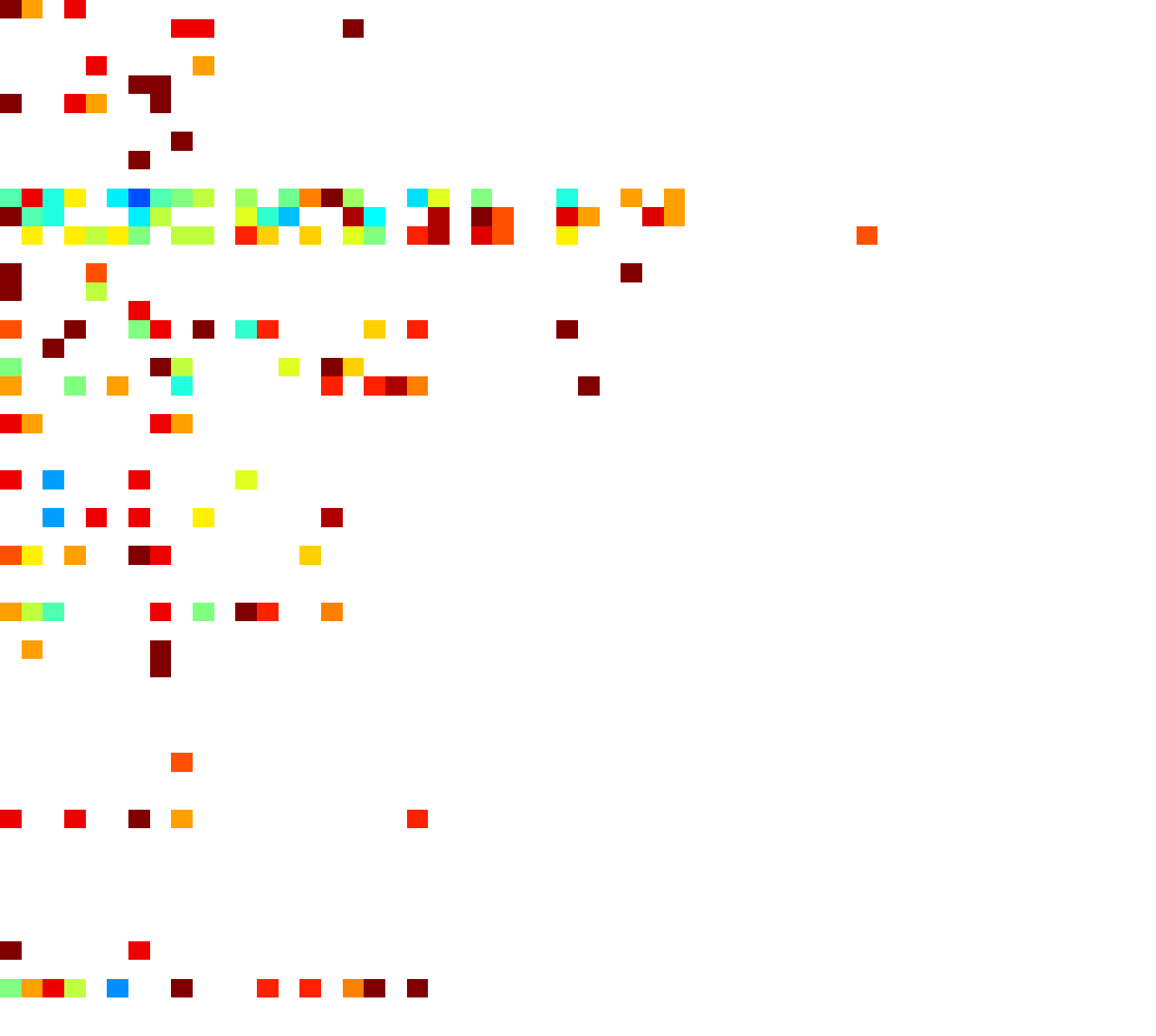}
  \end{subfigure}
    \begin{subfigure}[b]{0.18\textwidth}
    \includegraphics[width=\textwidth]{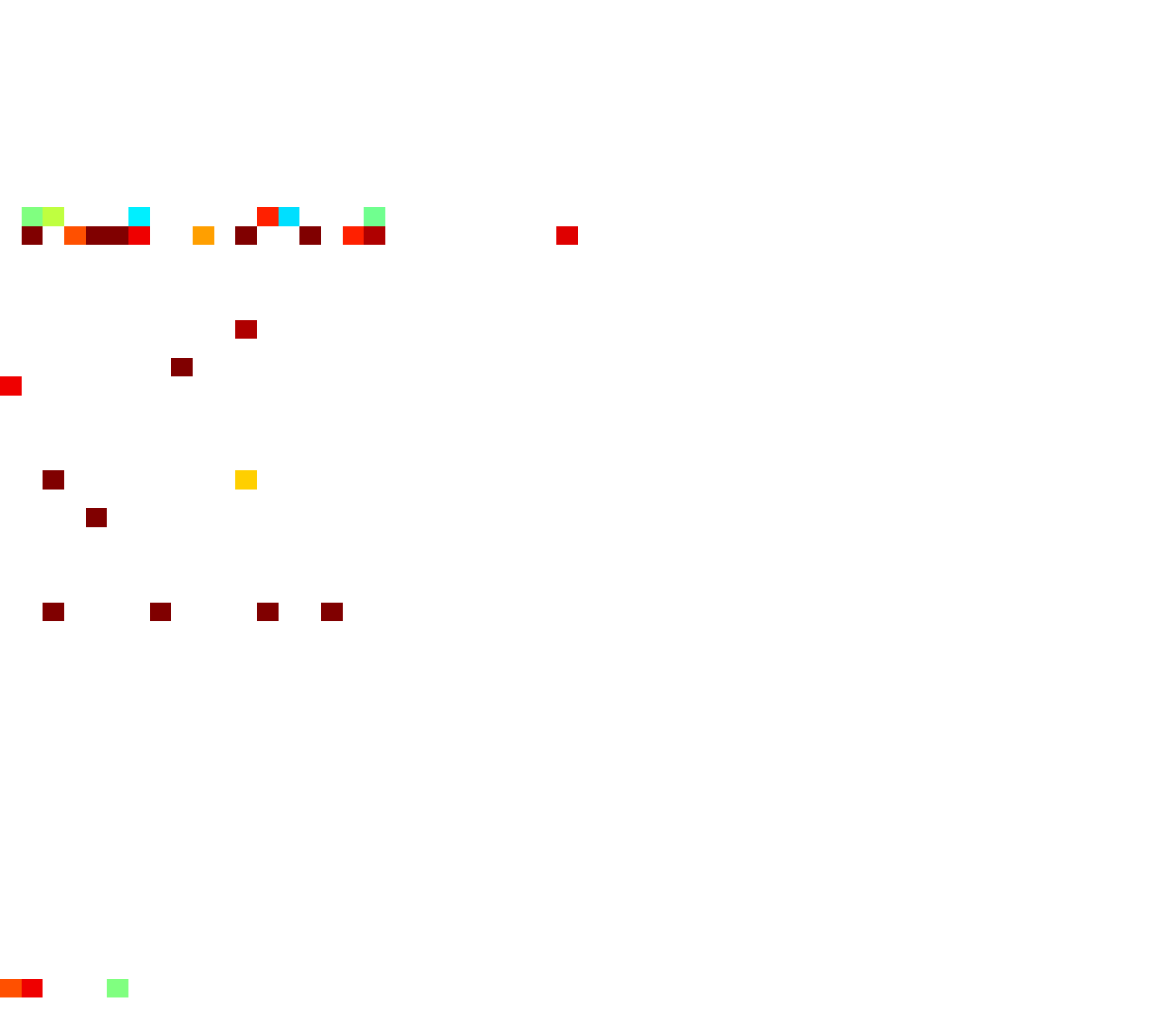}
  \end{subfigure}
  
  \begin{adjustbox}{angle=90}
  	\begin{subfigure}[b]{0.16\textwidth}
  		\hspace{2pt}
    	\caption{SAPEO-uf-ho}
  	\end{subfigure}
  \end{adjustbox}
  \begin{subfigure}[b]{0.18\textwidth}
    \includegraphics[width=\textwidth]{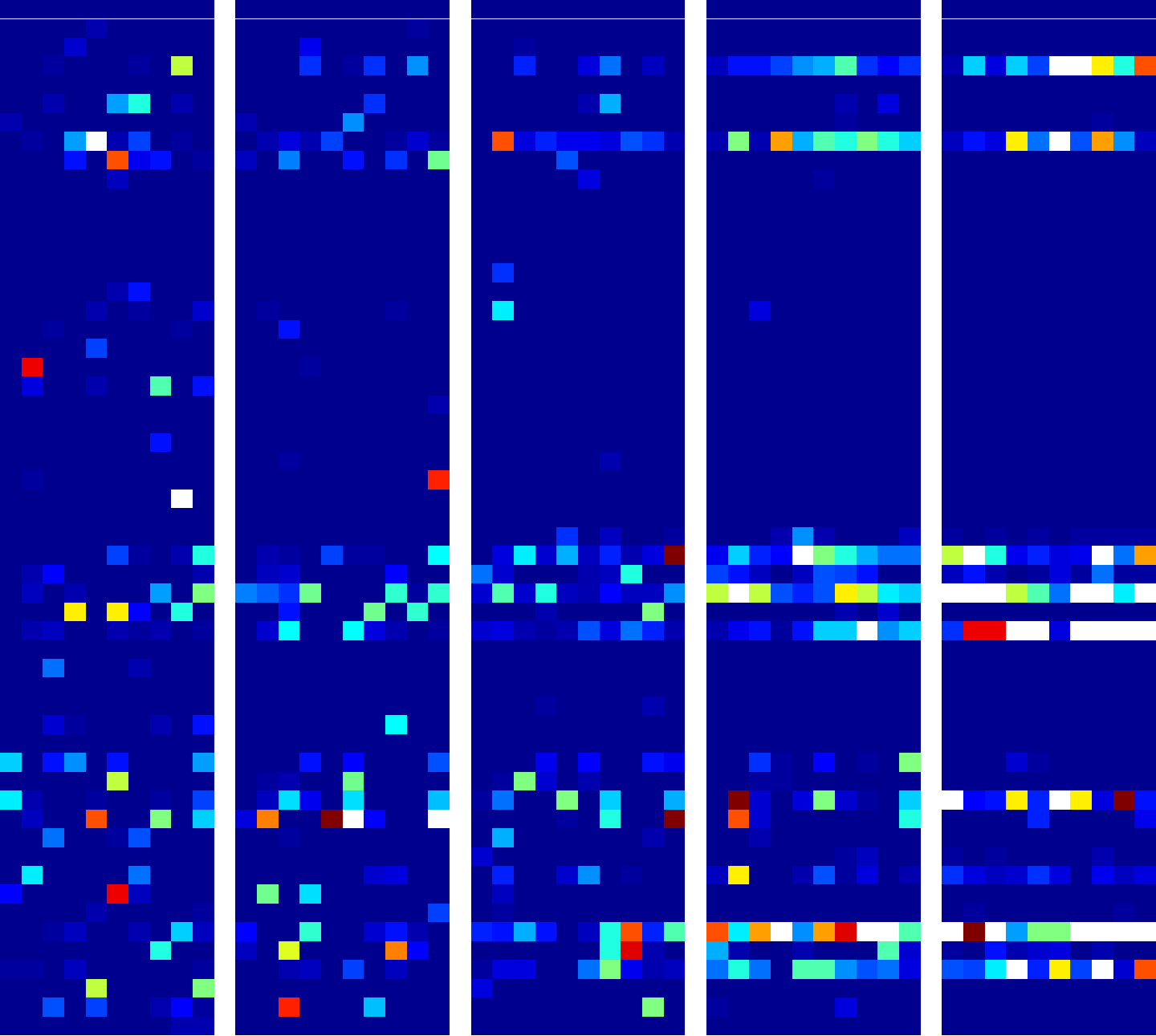}
  \end{subfigure}
  \begin{subfigure}[b]{0.18\textwidth}
    \includegraphics[width=\textwidth]{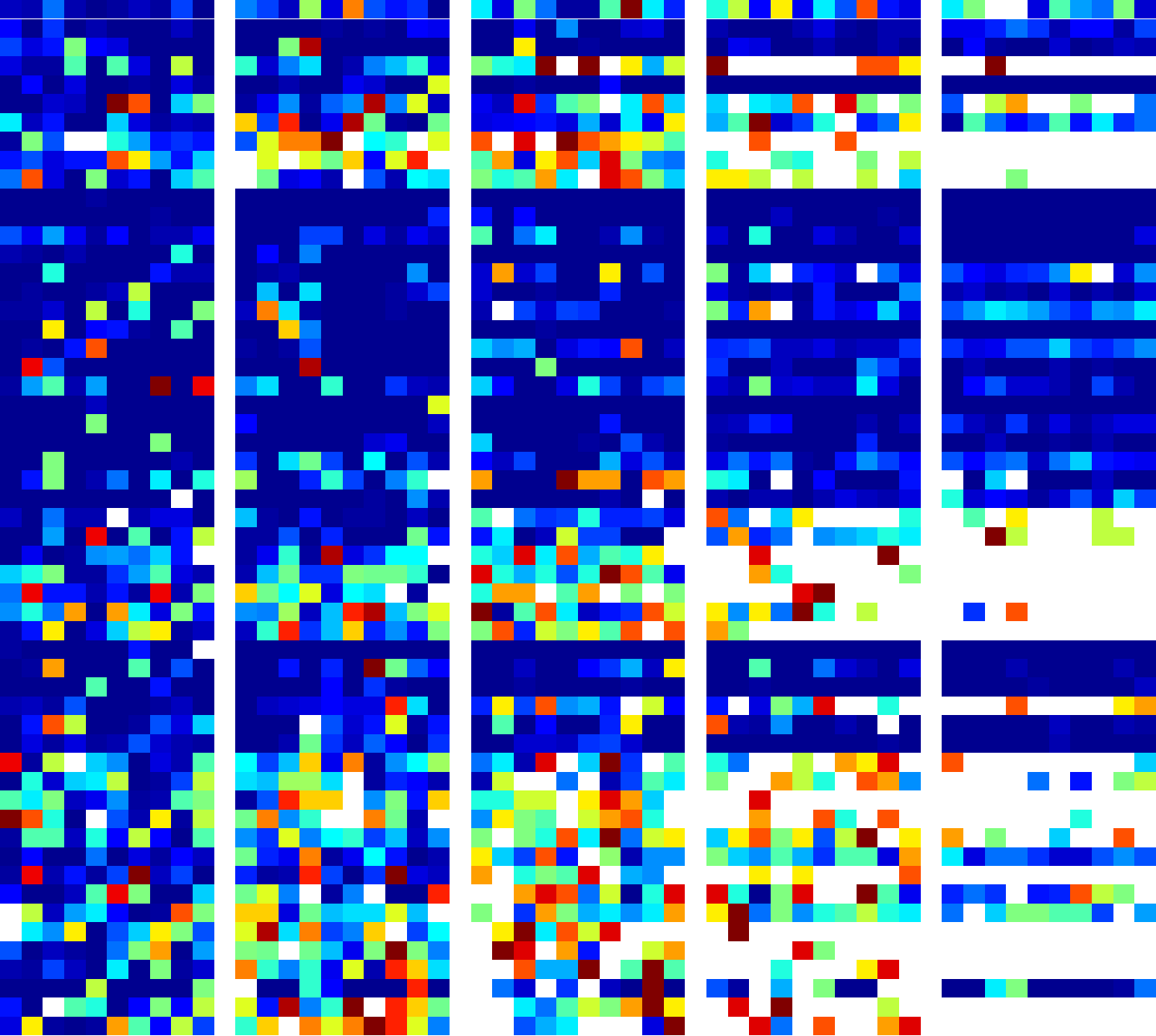}
  \end{subfigure}
    \begin{subfigure}[b]{0.18\textwidth}
    \includegraphics[width=\textwidth]{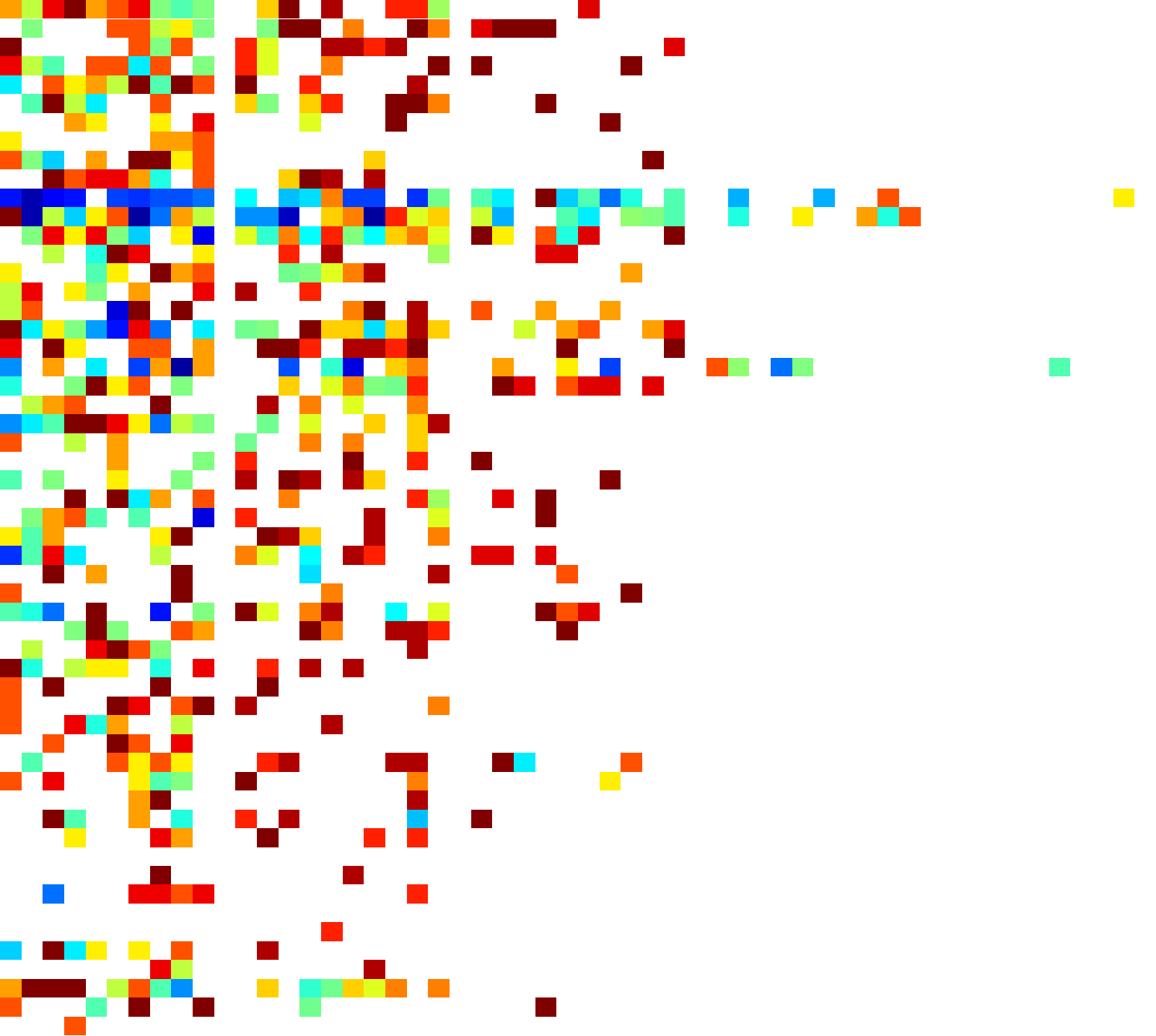}
  \end{subfigure}
    \begin{subfigure}[b]{0.18\textwidth}
    \includegraphics[width=\textwidth]{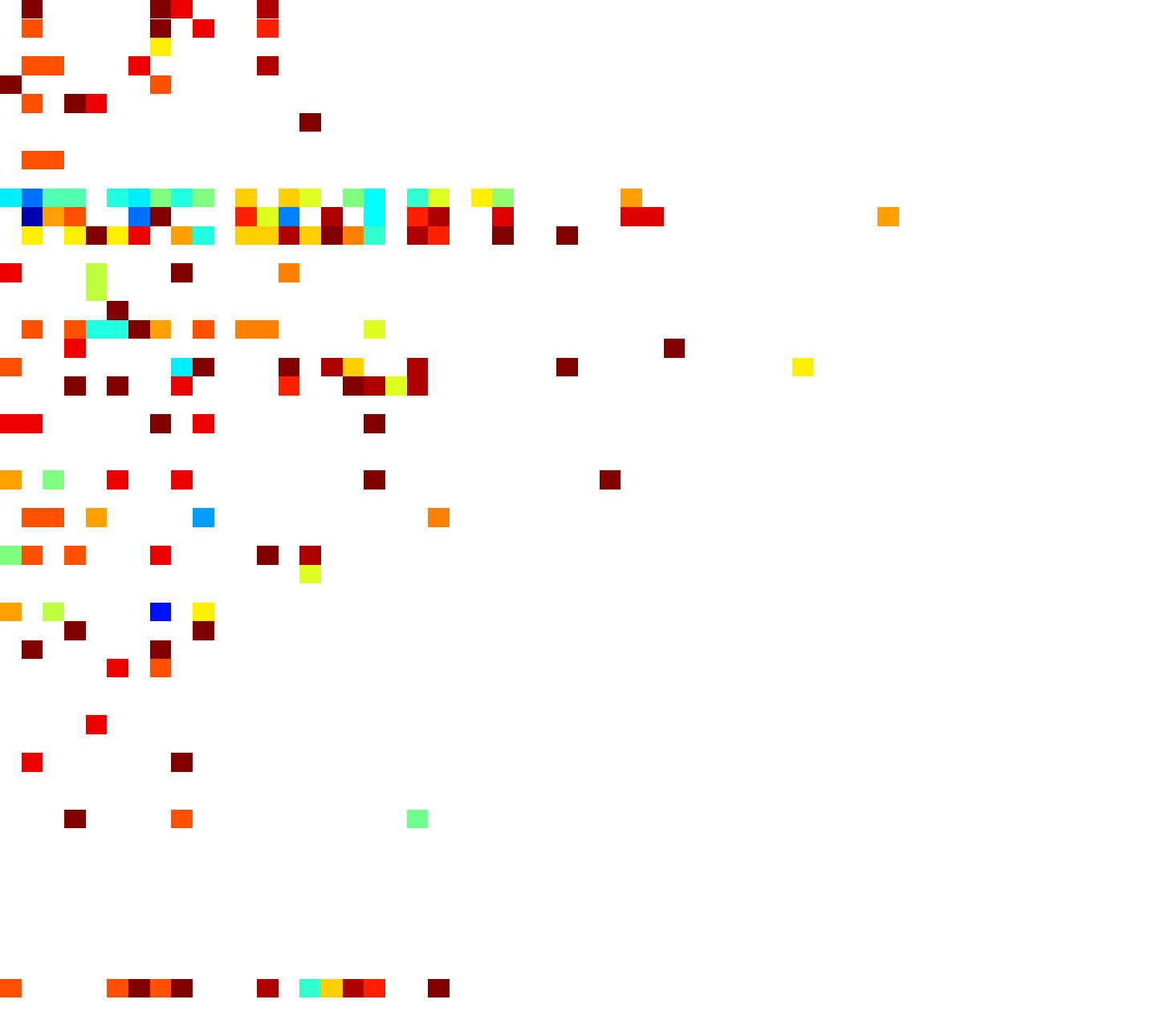}
  \end{subfigure}
    \begin{subfigure}[b]{0.18\textwidth}
    \includegraphics[width=\textwidth]{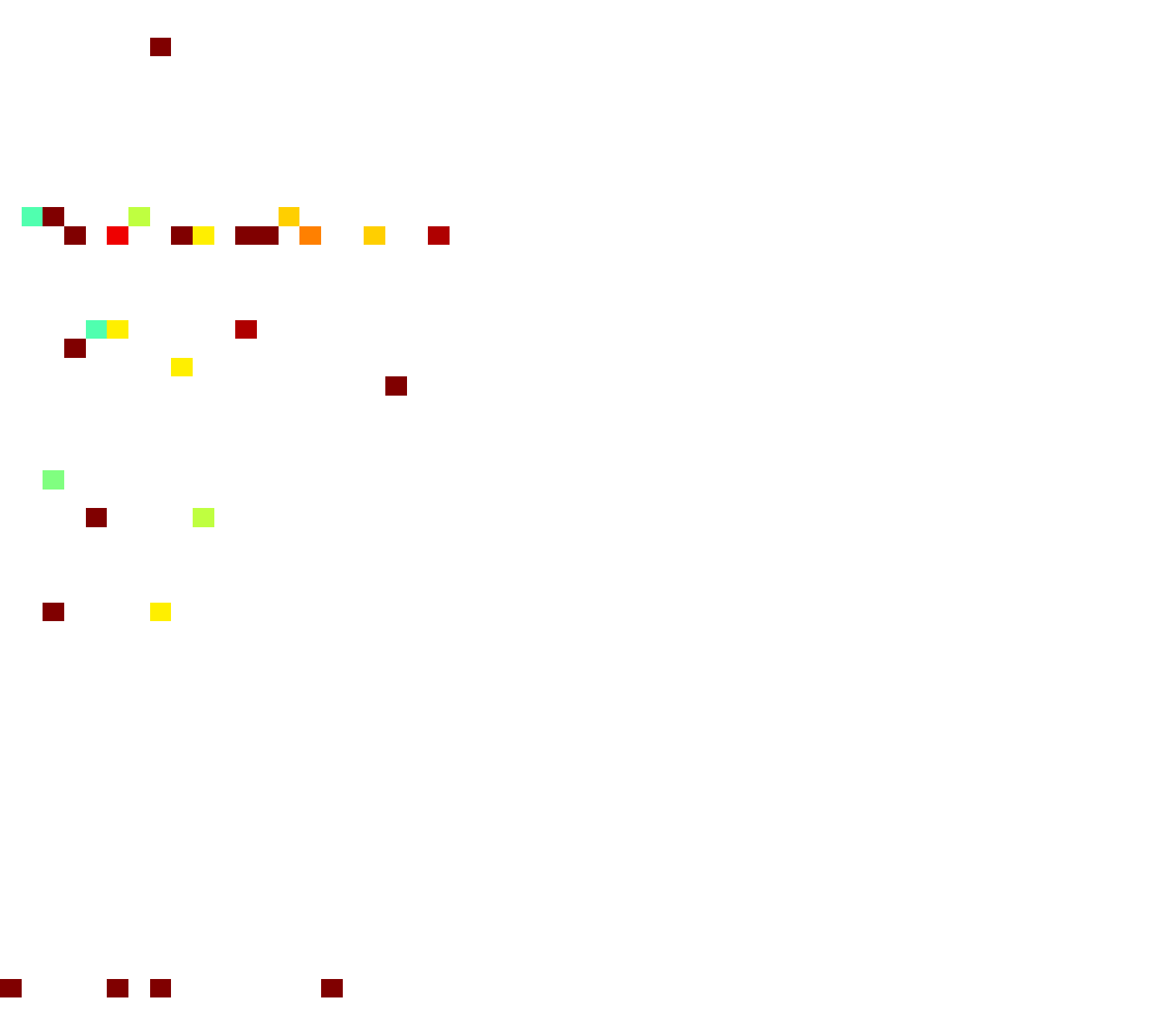}
  \end{subfigure}
  
  \begin{adjustbox}{angle=90}
  	\begin{subfigure}[b]{0.16\textwidth}
  		\hspace{2pt}
    	\caption{SA-SMS-o}
  	\end{subfigure}
  \end{adjustbox}
  \begin{subfigure}[b]{0.18\textwidth}
    \includegraphics[width=\textwidth]{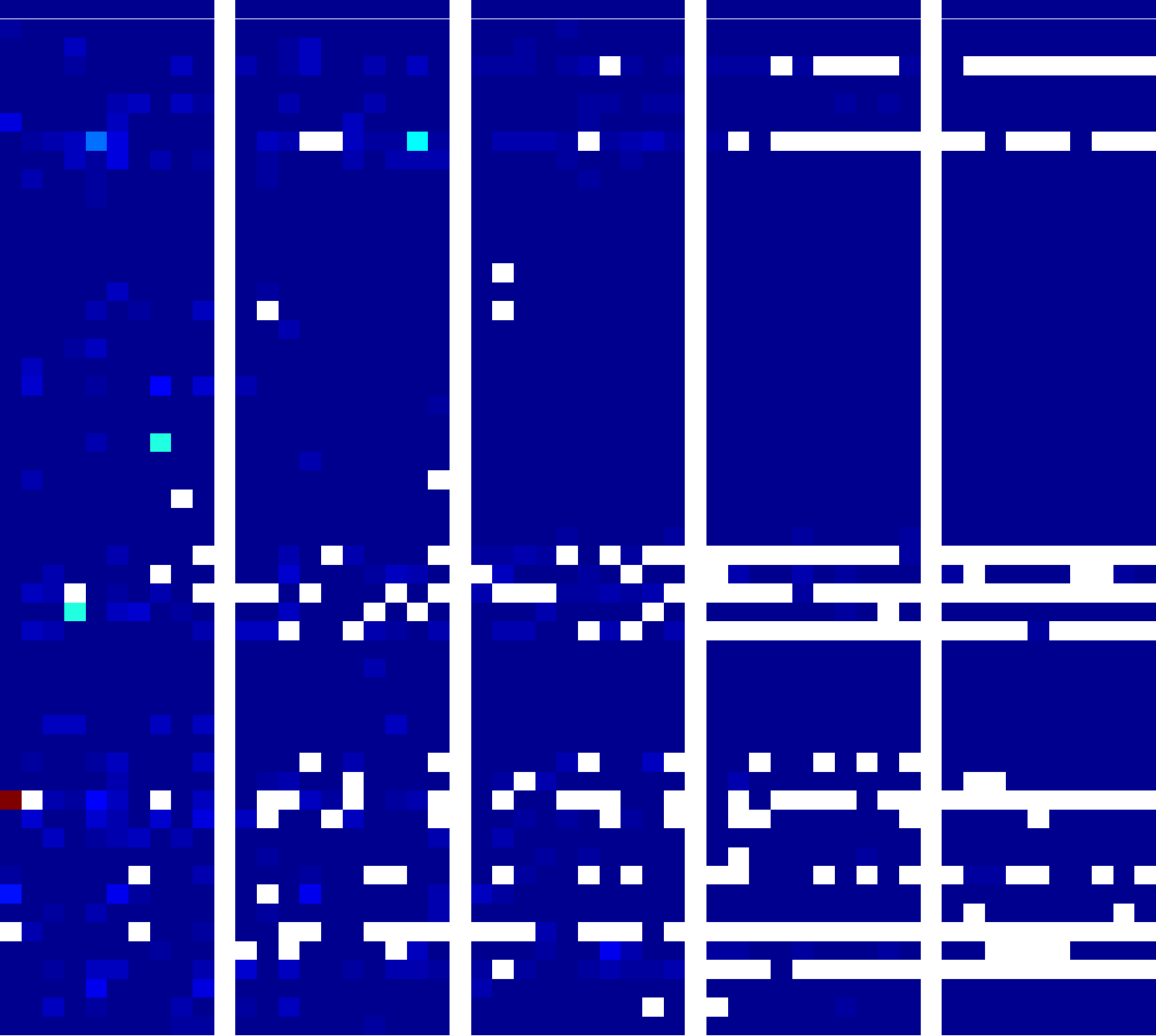}
  \end{subfigure}
  \begin{subfigure}[b]{0.18\textwidth}
    \includegraphics[width=\textwidth]{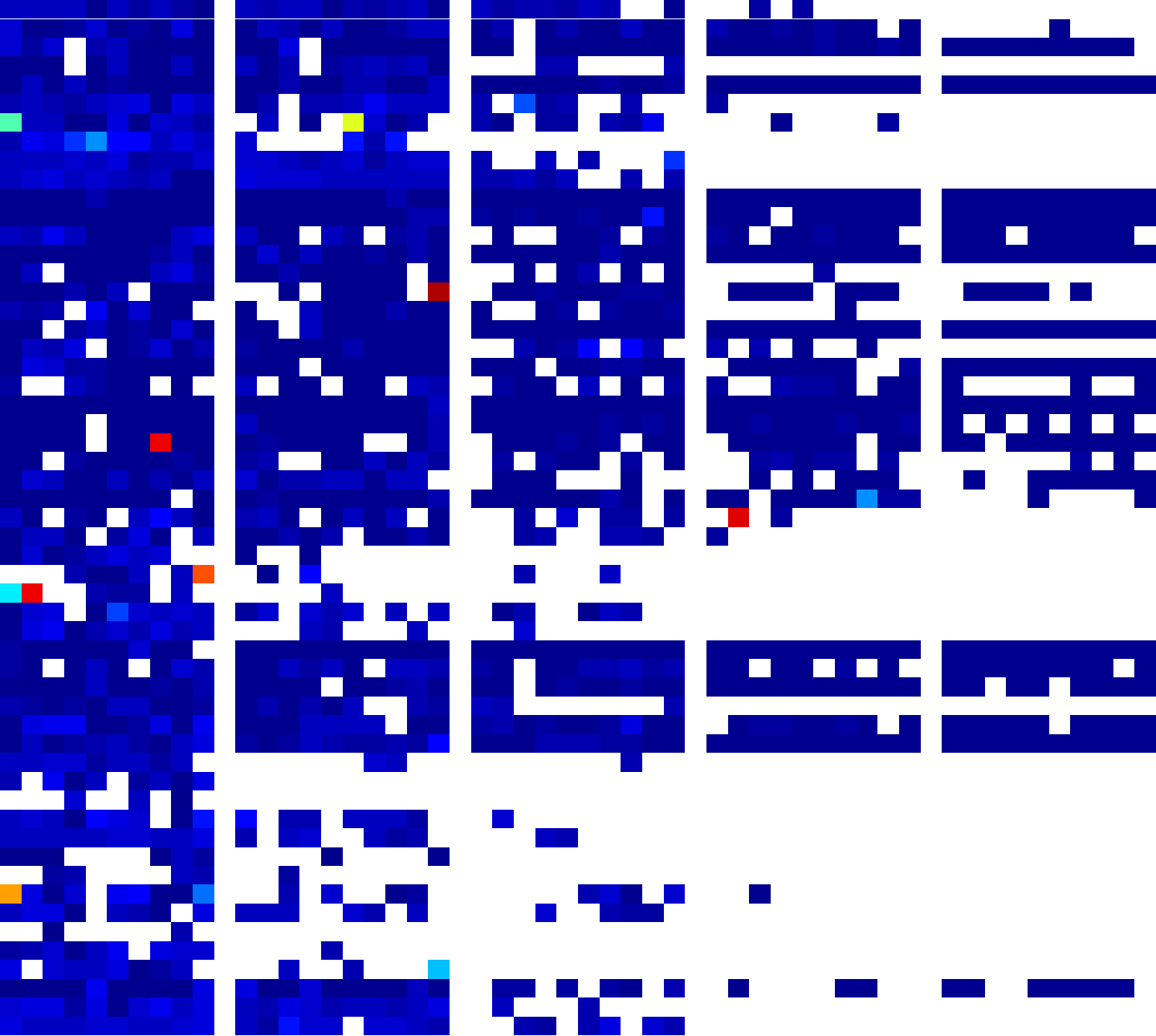}
  \end{subfigure}
    \begin{subfigure}[b]{0.18\textwidth}
    \includegraphics[width=\textwidth]{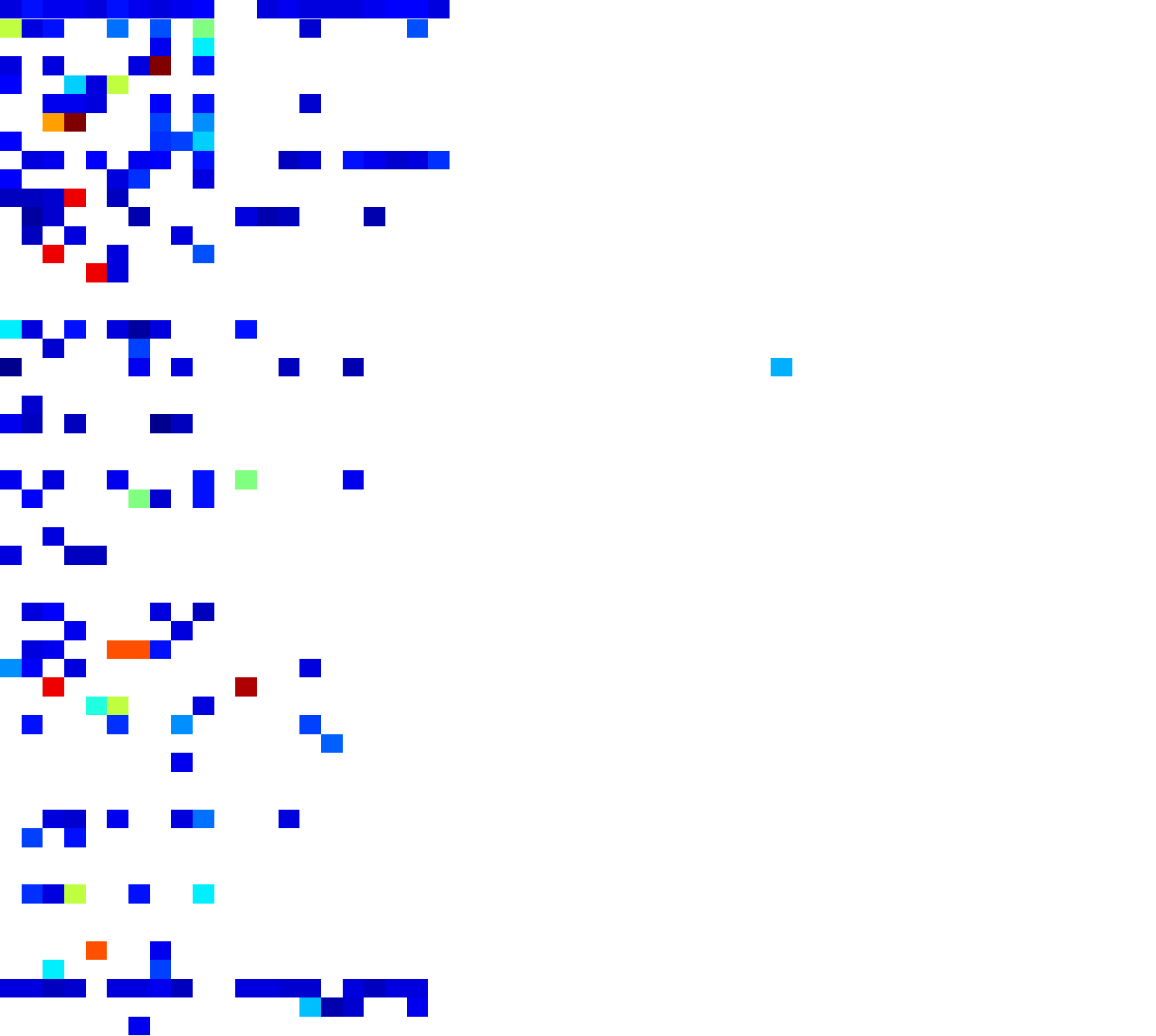}
  \end{subfigure}
    \begin{subfigure}[b]{0.18\textwidth}
    \includegraphics[width=\textwidth]{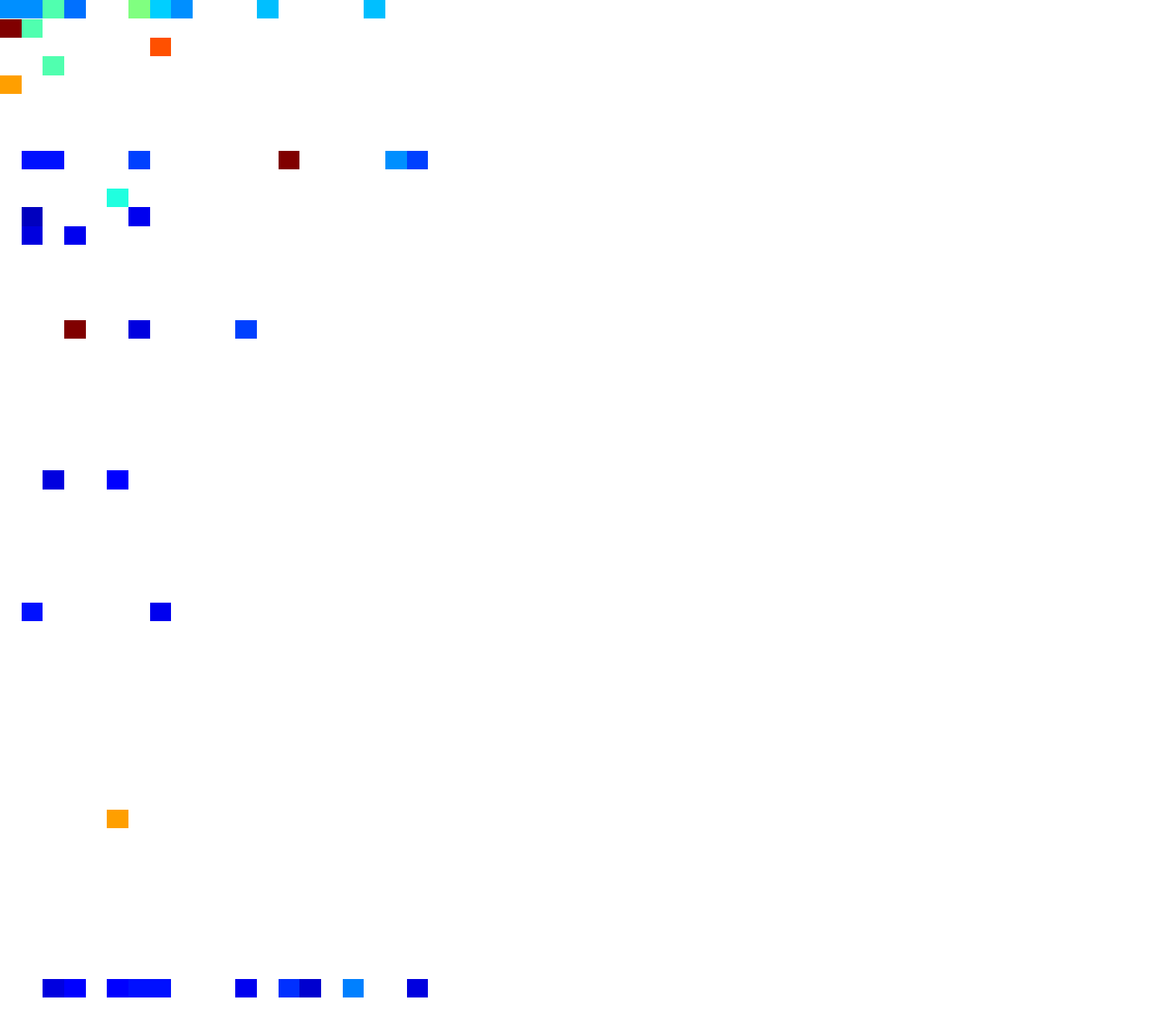}
  \end{subfigure}
    \begin{subfigure}[b]{0.18\textwidth}
    \includegraphics[width=\textwidth]{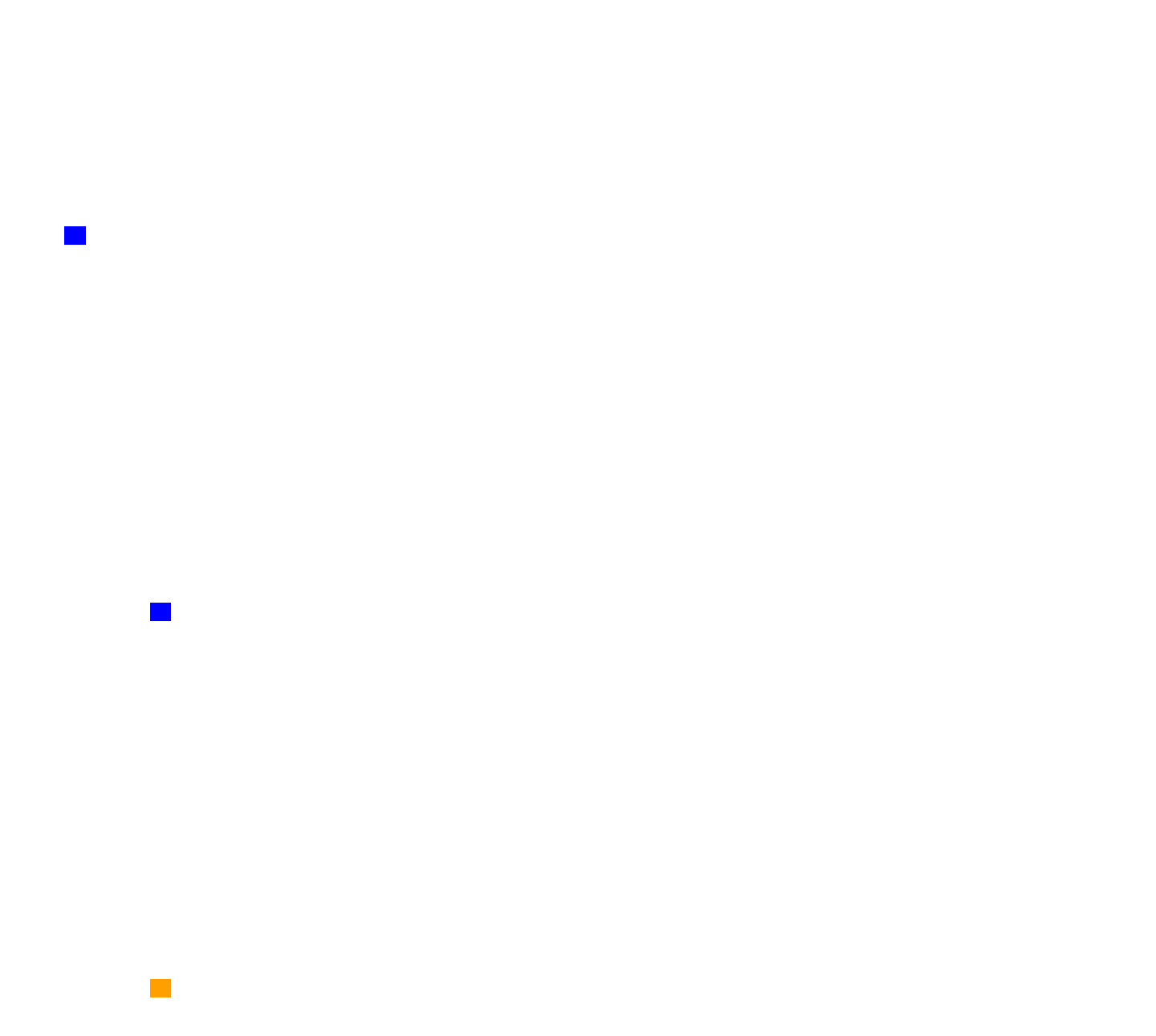}
  \end{subfigure}
  
  \begin{adjustbox}{angle=90}
  	\begin{subfigure}[b]{0.16\textwidth}
  		\hspace{2pt}
    	\caption{SA-SMS-p}
  	\end{subfigure}
  \end{adjustbox}
  \begin{subfigure}[b]{0.18\textwidth}
    \includegraphics[width=\textwidth]{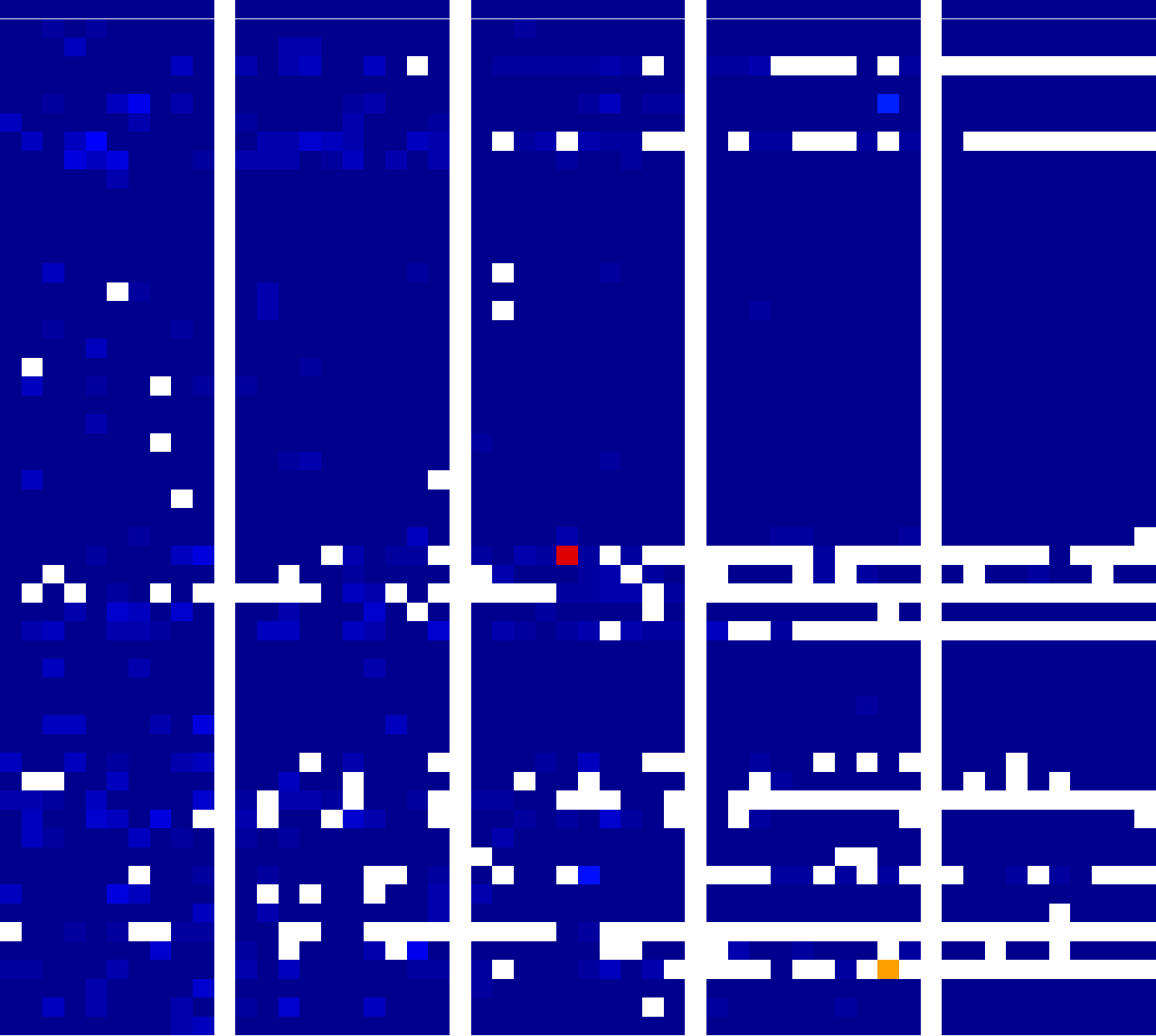}
  \end{subfigure}
  \begin{subfigure}[b]{0.18\textwidth}
    \includegraphics[width=\textwidth]{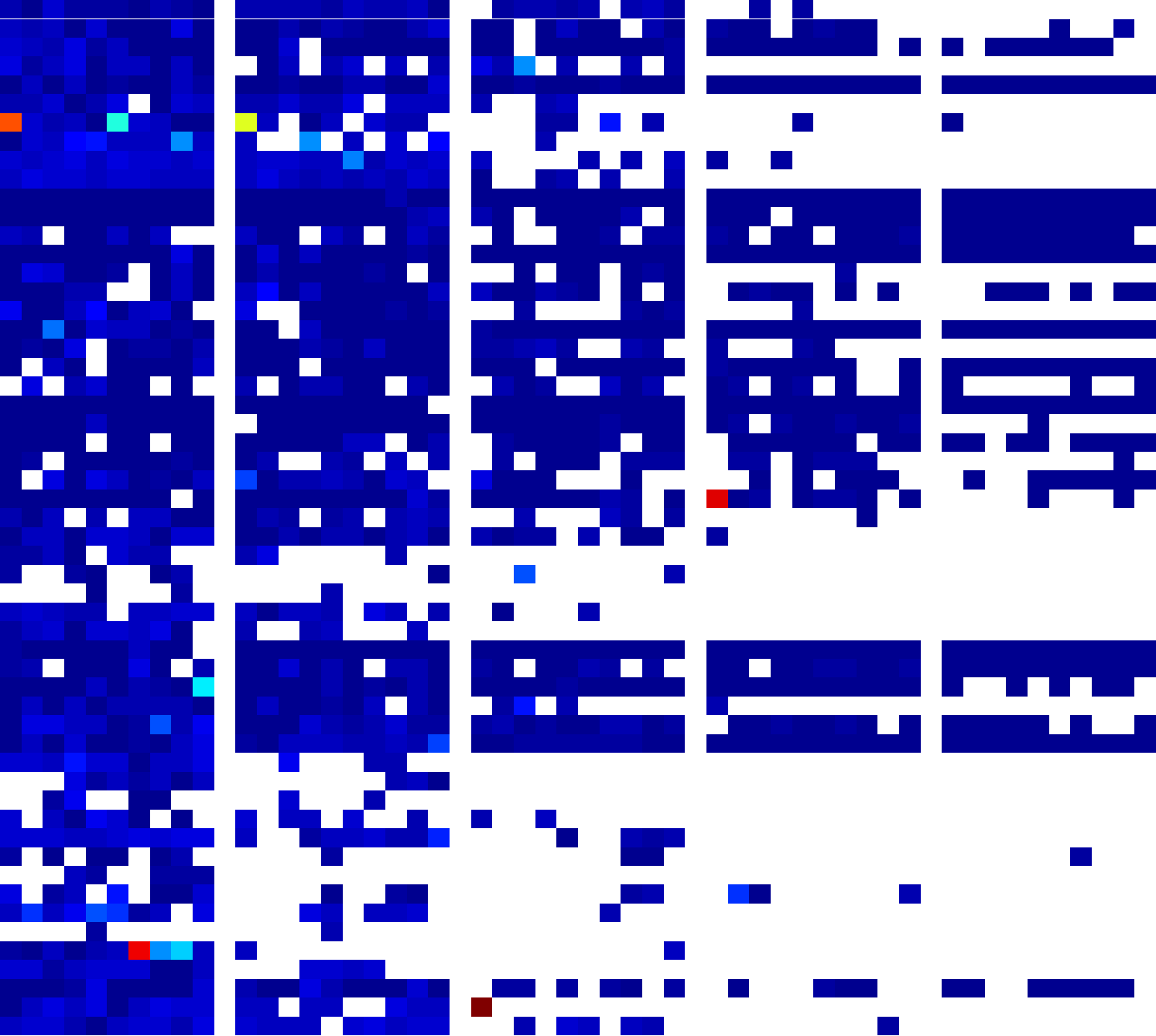}
  \end{subfigure}
    \begin{subfigure}[b]{0.18\textwidth}
    \includegraphics[width=\textwidth]{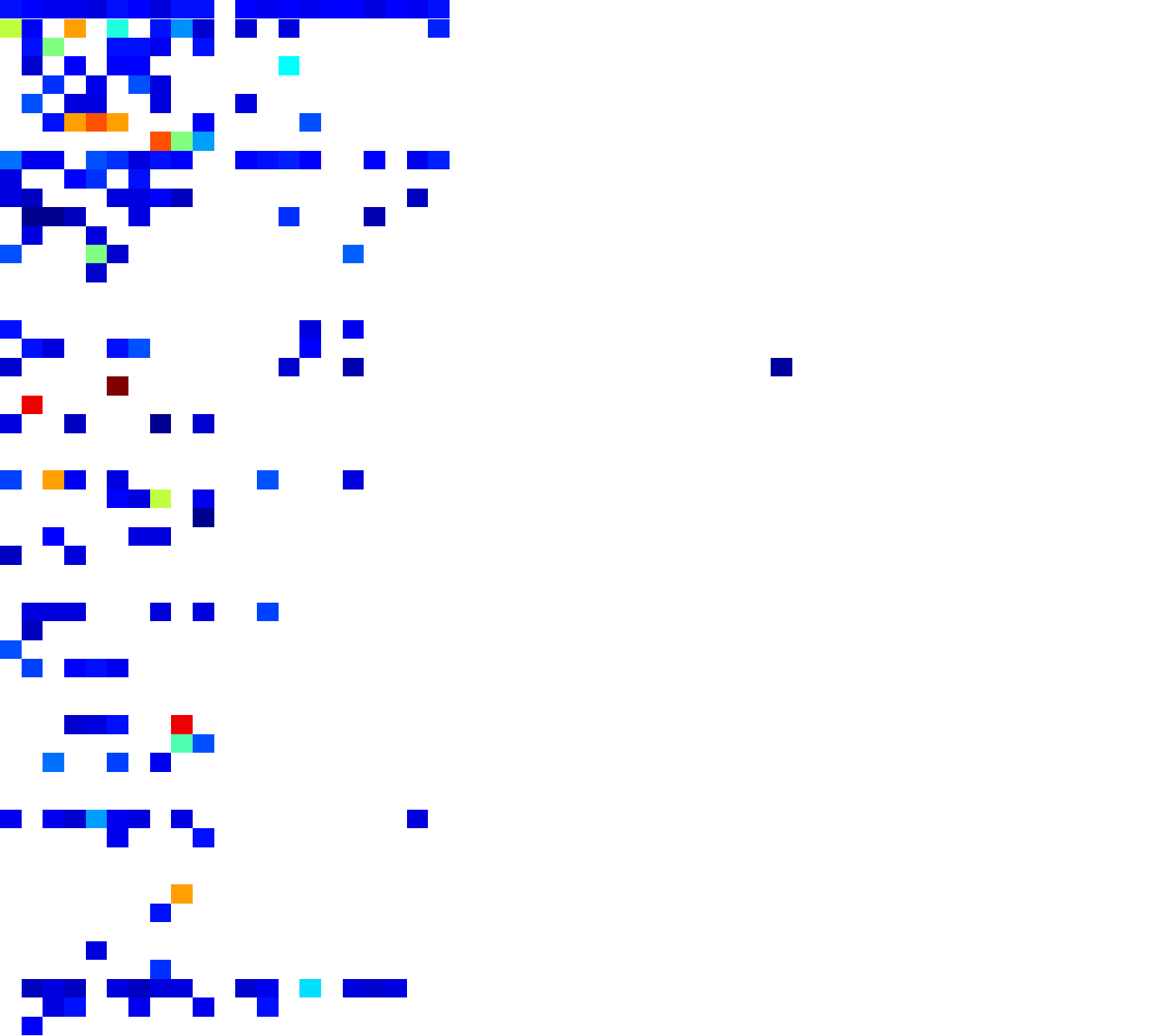}
  \end{subfigure}
    \begin{subfigure}[b]{0.18\textwidth}
    \includegraphics[width=\textwidth]{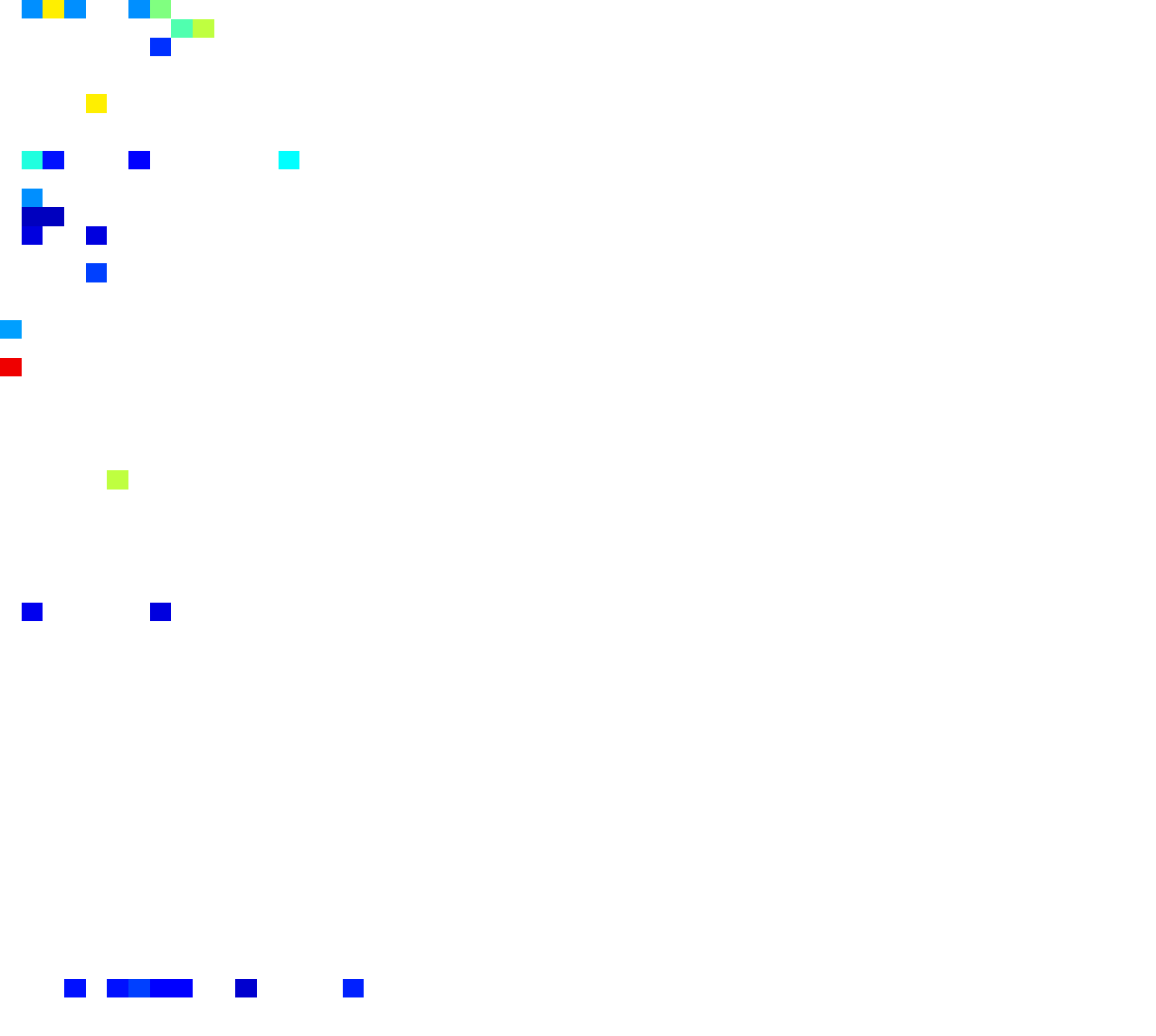}
  \end{subfigure}
    \begin{subfigure}[b]{0.18\textwidth}
    \includegraphics[width=\textwidth]{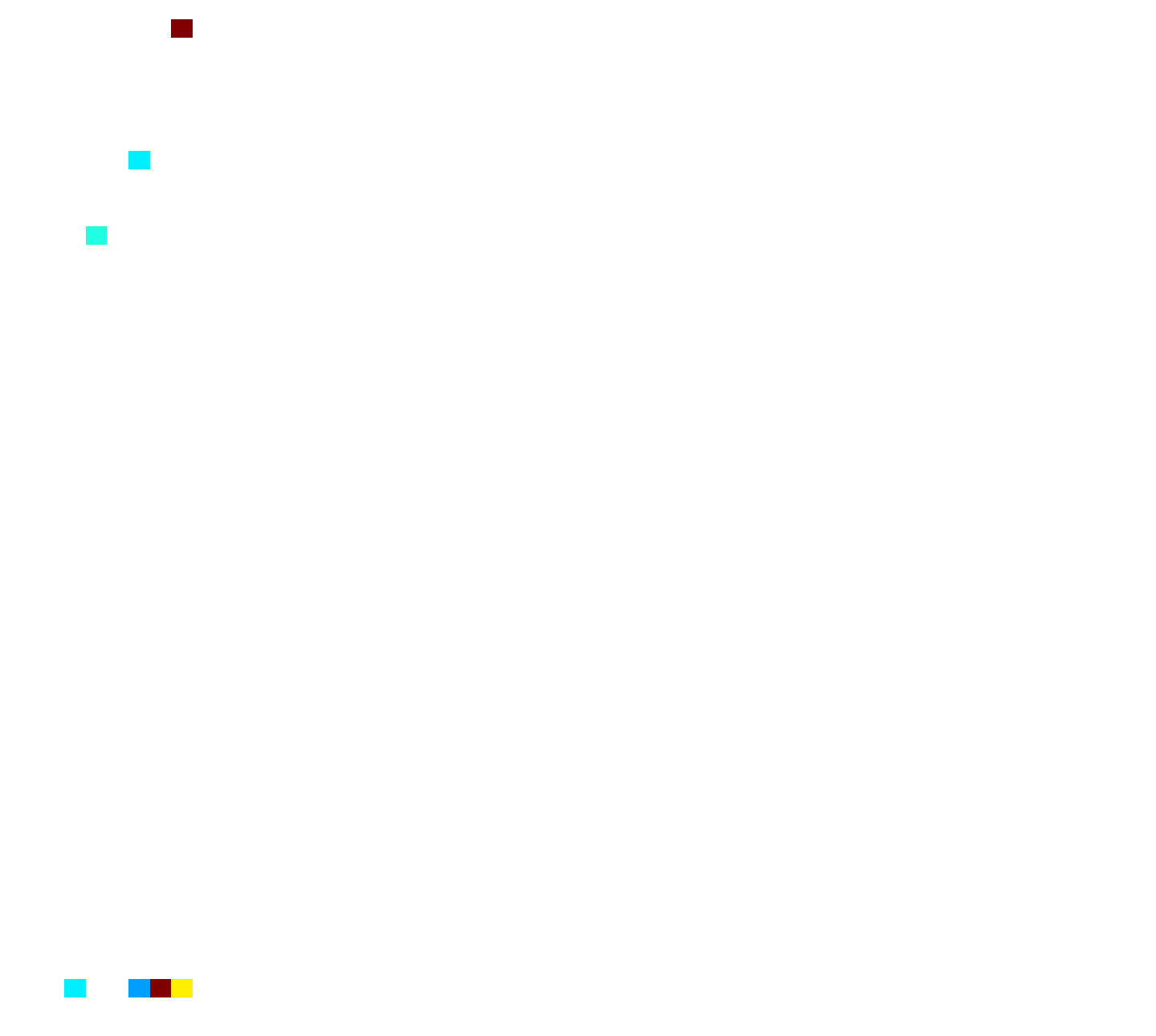}
  \end{subfigure}
  
  \begin{adjustbox}{angle=90}
  	\begin{subfigure}[b]{0.16\textwidth}
  		\hspace{2pt}
    	\caption{SMS-EMOA}
  	\end{subfigure}
  \end{adjustbox}
  \begin{subfigure}[b]{0.18\textwidth}
    \includegraphics[width=\textwidth]{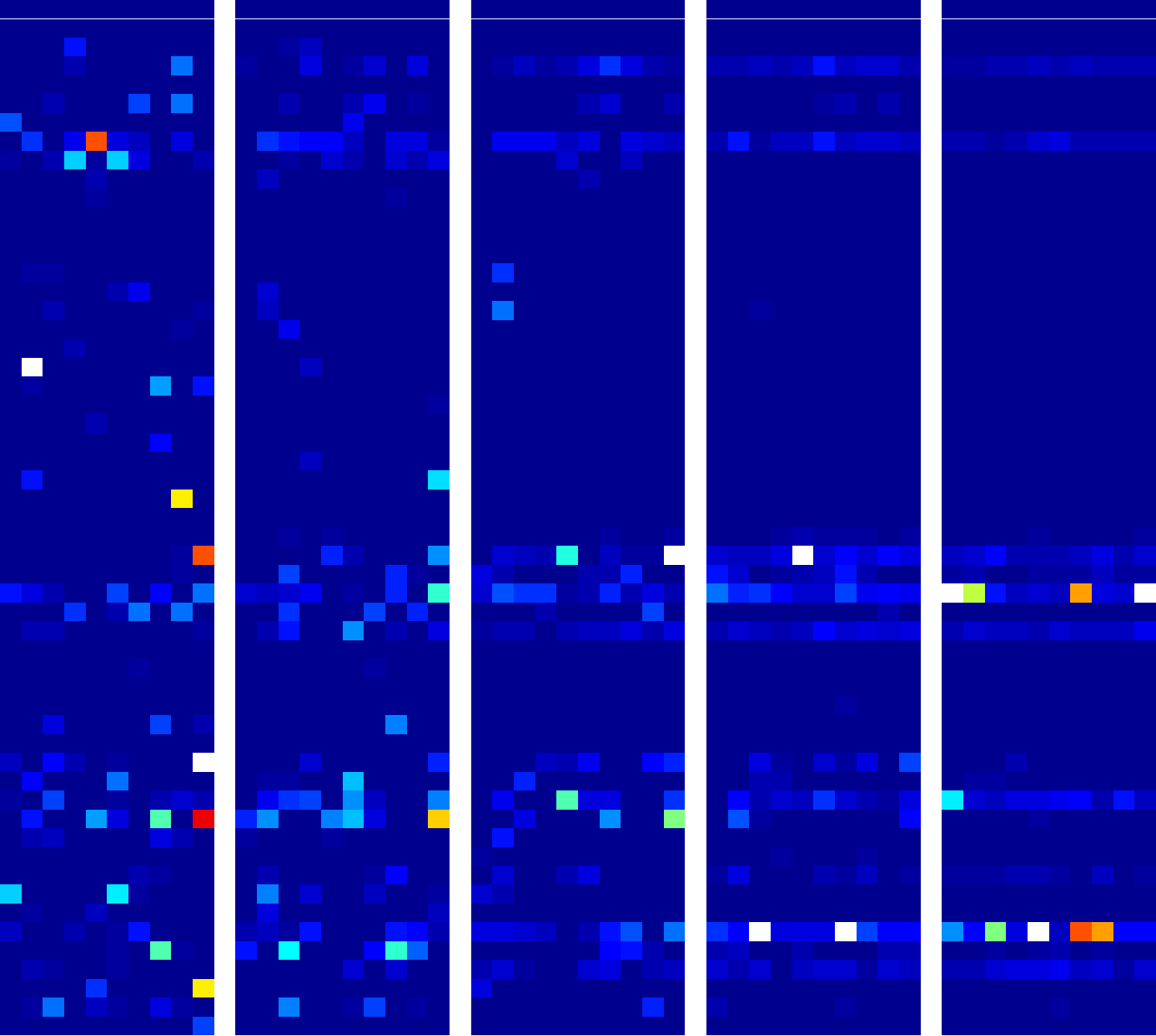}
  \end{subfigure}
  \begin{subfigure}[b]{0.18\textwidth}
    \includegraphics[width=\textwidth]{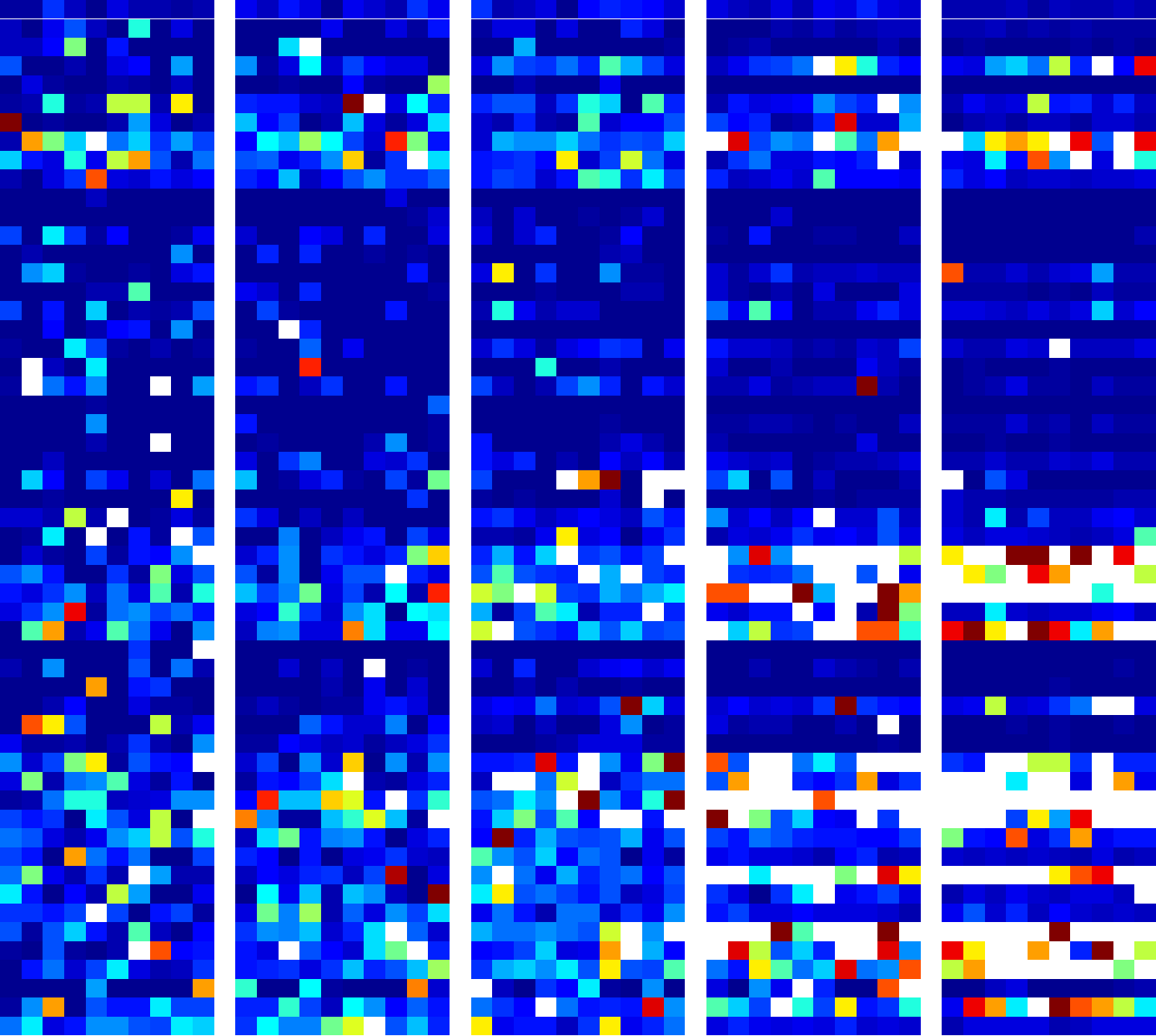}
  \end{subfigure}
    \begin{subfigure}[b]{0.18\textwidth}
    \includegraphics[width=\textwidth]{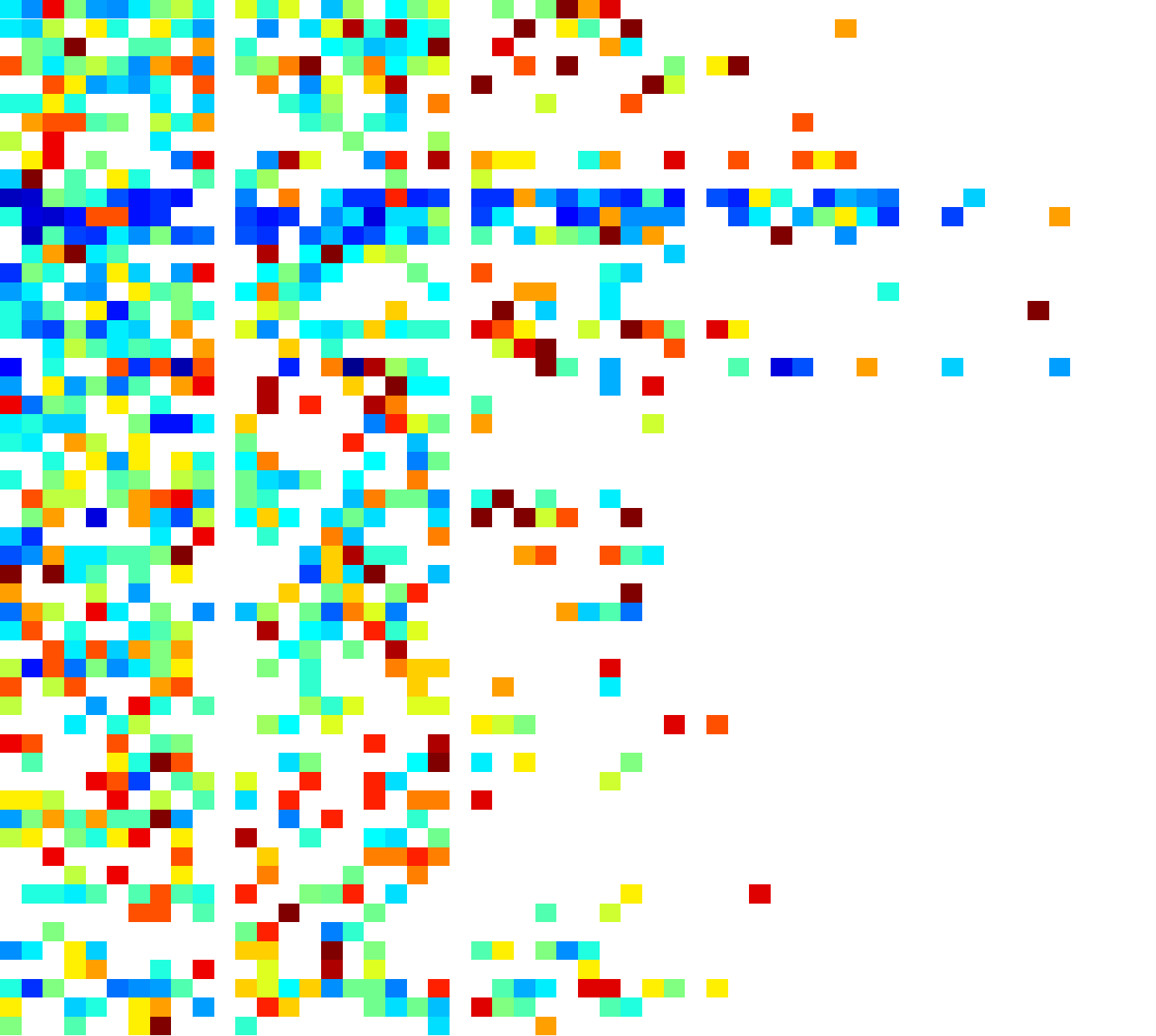}
  \end{subfigure}
    \begin{subfigure}[b]{0.18\textwidth}
    \includegraphics[width=\textwidth]{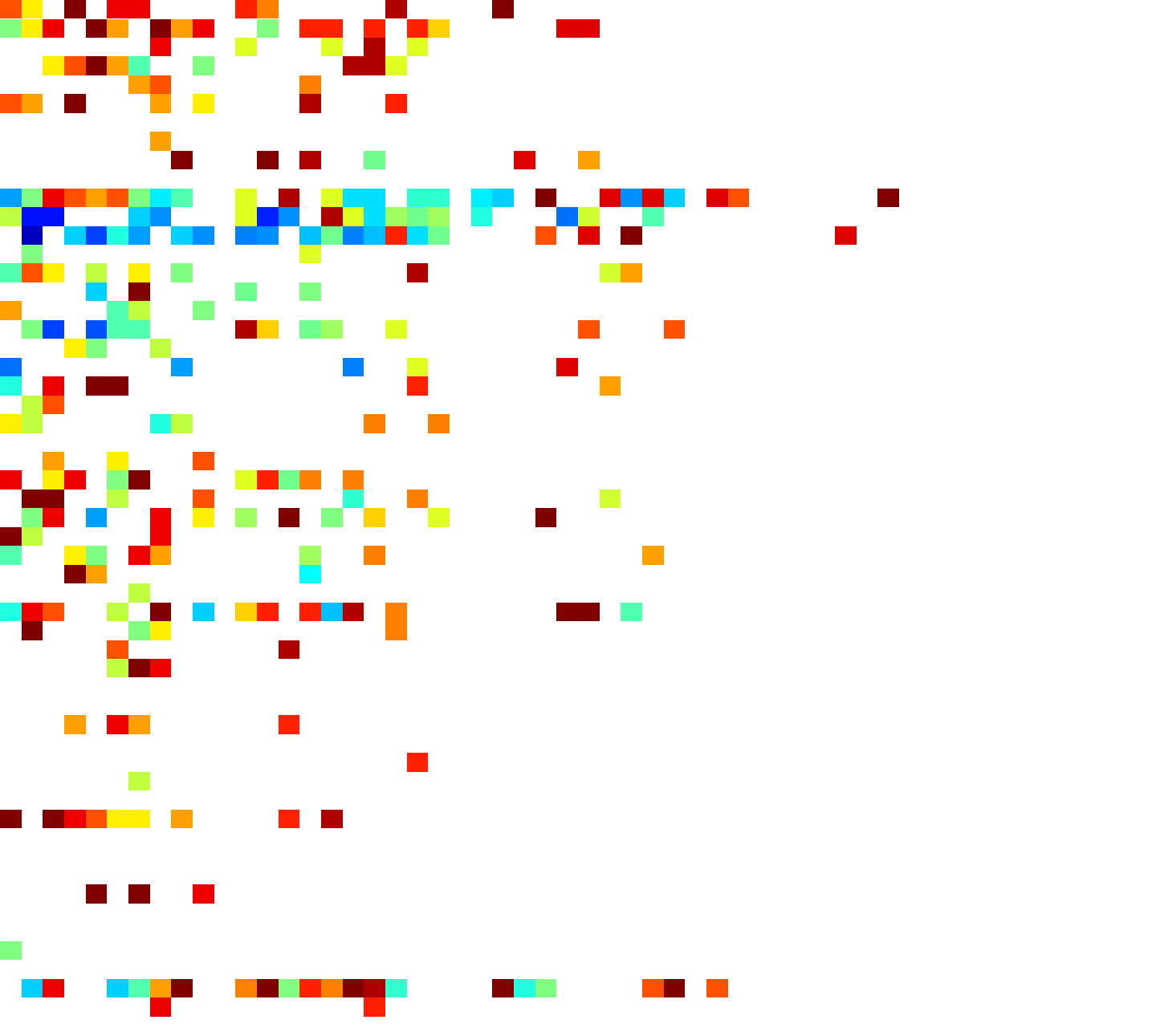}
  \end{subfigure}
    \begin{subfigure}[b]{0.18\textwidth}
    \includegraphics[width=\textwidth]{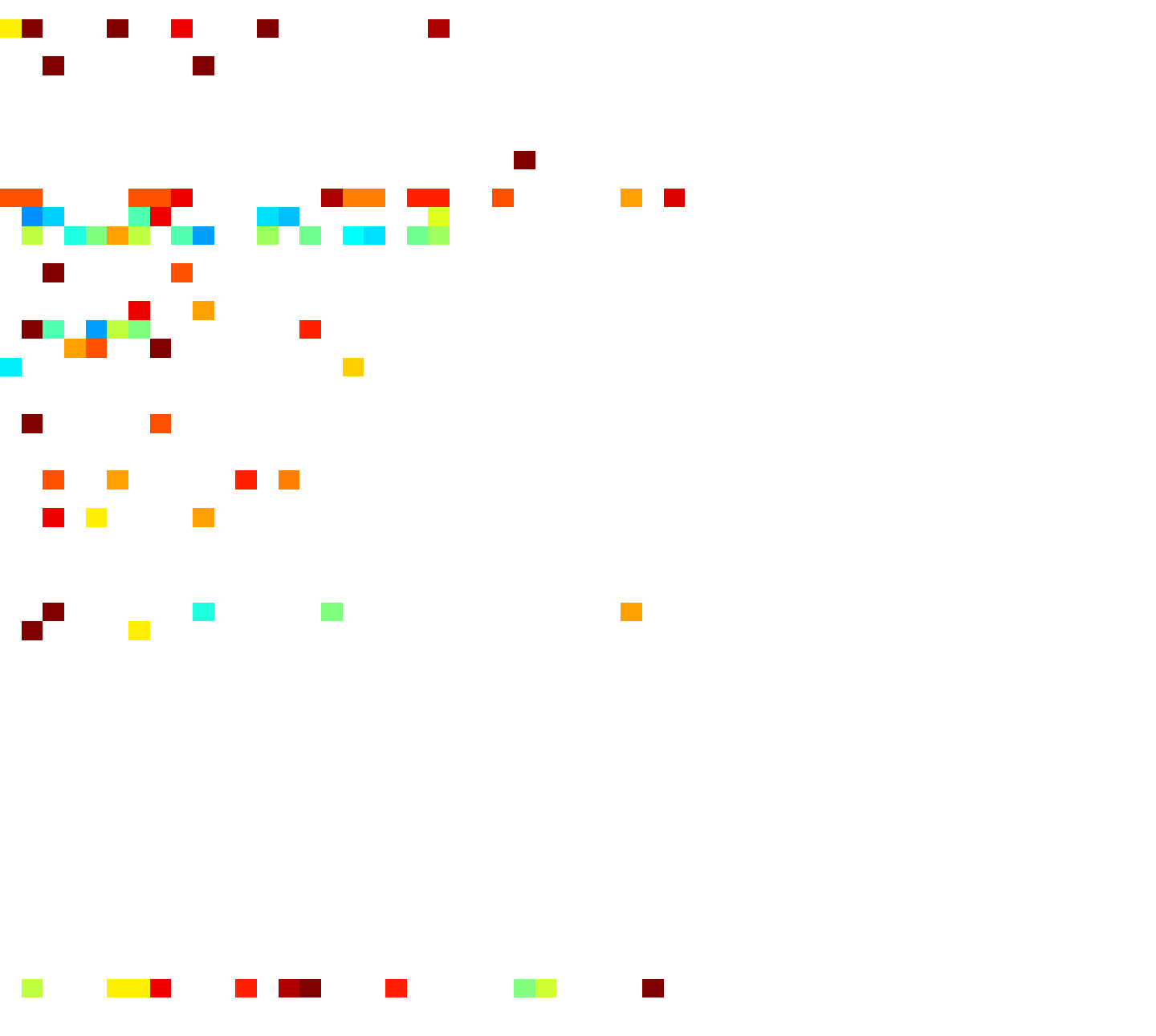}
  \end{subfigure}
  
  \caption{Heatmaps visualising target performances for all algorithms (rows) across multiple targets (columns). Refer to figure \ref{fig:sapeobig} for a detailed explanation.}
  \label{fig:mega}
\end{figure}